\documentclass{article}

\usepackage[main,final]{neurips_2025}

\newcommand{\spara}[1]{\smallskip\noindent{\bf #1}}
\usepackage{subcaption} % 加载子图标题包

\usepackage{bm}
\AtBeginDocument{%
  }
\usepackage[linesnumbered,ruled,vlined]{algorithm2e}
\usepackage{todonotes}
\usepackage{multirow}
\usepackage{graphicx}
\usepackage{enumitem}
\usepackage{amsmath}
\usepackage{amsthm}
\usepackage{algpseudocode}
\usepackage{wrapfig}
\usepackage{booktabs}
\usepackage{caption}
\usepackage{lipsum}  % 示例文字
\usepackage[utf8]{inputenc} % allow utf-8 input
\usepackage[T1]{fontenc}    % use 8-bit T1 fonts
\usepackage{hyperref}       % hyperlinks
\usepackage{url}            % simple URL typesetting
\usepackage{booktabs}       % professional-quality tables
\usepackage{amsfonts}       % blackboard math symbols
\usepackage{nicefrac}       % compact symbols for 1/2, etc.
\usepackage{microtype}      % microtypography
\usepackage{xcolor}         % colors
\usepackage{float}
\usepackage{booktabs}  % 提供 \midrule 等命令
\usepackage{lipsum}    % 用于生成示例文本
\usepackage{amsthm}
\usepackage{amsmath}
\newtheorem{myDef}{\textbf{Definition}}

\AtBeginDocument{%
  }
    
\title{Uncertain Knowledge Graph Completion via Semi-Supervised Confidence Distribution Learning}

\author{
Tianxing Wu\textsuperscript{1,2,}\thanks{Corresponding authors.}\quad Shutong Zhu\textsuperscript{1}\quad Jingting Wang\textsuperscript{1}\quad \textbf{Ning Xu\textsuperscript{1,2,}\footnotemark[1]}\\\quad \textbf{Guilin Qi\textsuperscript{1,2}}\quad \textbf{Haofen Wang\textsuperscript{3}}\\
\textsuperscript{1}School of Computer Science and Engineering, Southeast University, China\\ 
\textsuperscript{2}Key Laboratory of New Generation Artificial Intelligence Technology and\\ its Interdisciplinary Applications (Southeast University), Ministry of Education, China\\
\textsuperscript{3}College of Design and Innovation, Tongji University, China\\
\texttt{\{tianxingwu, shutong\_zhu, xning\}@seu.edu.cn}
}

\begin{document}

\maketitle

\begin{abstract}
Uncertain knowledge graphs (UKGs) associate each triple with a confidence score to provide more precise knowledge representations. Recently, since real-world UKGs suffer from the incompleteness, uncertain knowledge graph (UKG) completion attracts more attention, aiming to complete missing triples and confidences. Current studies attempt to learn UKG embeddings to solve this problem, but they neglect the extremely imbalanced distributions of triple confidences. This causes that the learnt embeddings are insufficient to high-quality UKG completion. Thus, in this paper, to address the above issue, we propose a new \underline{\textbf{s}}emi-\underline{\textbf{s}}upervised \underline{\textbf{C}}onfidence \underline{\textbf{D}}istribution \underline{\textbf{L}}earning (\textbf{ssCDL}) method for UKG completion, where each triple confidence is transformed into a confidence distribution to introduce more supervision information of different confidences to reinforce the embedding learning process. ssCDL iteratively learns UKG embedding by relational learning on labeled data (i.e., existing triples with confidences) and unlabeled data with pseudo labels (i.e., unseen triples with the generated confidences), which are predicted by meta-learning to augment the training data and rebalance the distribution of triple confidences. Experiments on two UKG datasets demonstrate that ssCDL consistently outperforms state-of-the-art baselines in different evaluation metrics.
\end{abstract}

\section{Introduction}

Knowledge Graphs (KGs) are usually defined as multi-relational graphs describing knowledge with deterministic triples, each of which is in the form of (\textit{subject}, \textit{predicate}, \textit{object}), e.g., (\textit{Michael Jordan}, \textit{Nationality}, \textit{U.S.}). Such a kind of structured knowledge has supported many applications, including question answering~\cite{lan2021}, semantic search~\cite{galhotra2020semantic}, decision-making systems~\cite{you2023knowledge}, and etc. Recently, uncertain KGs (UKGs), such as NELL~\cite{carlson2010toward}, ConceptNet~\cite{speer2017conceptnet}, and Probase~\cite{wu2012probase,ji2019microsoft}, have received much more attention. UKGs measure the uncertainty of knowledge by associating each triple with a confidence score, which also denotes the likelihood of that triple to be true. Such a setting benefits to precise knowledge representation and reasoning in the real world.

\begin{figure}[t]
  \centering
  \begin{subfigure}[b]{0.32\textwidth}  % 调整为0.3或稍小
    \includegraphics[width=\textwidth]{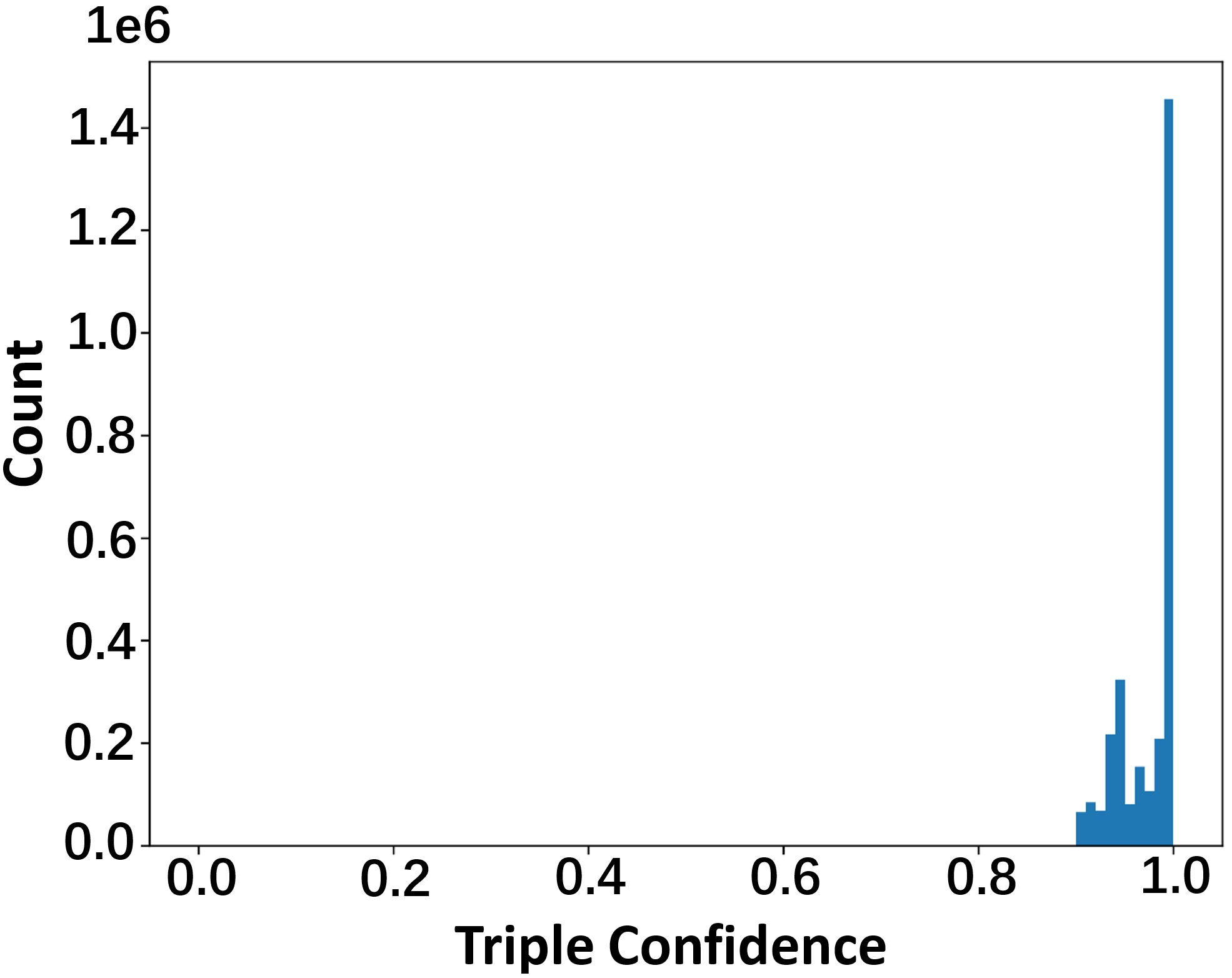}
     %\vspace{-1mm}
    \caption{}
    \label{fig:exp_conf_nell}
  \end{subfigure}
  \hfill  % 使用\hfill填充水平空间
  \begin{subfigure}[b]{0.64\textwidth}
    \includegraphics[width=\textwidth]{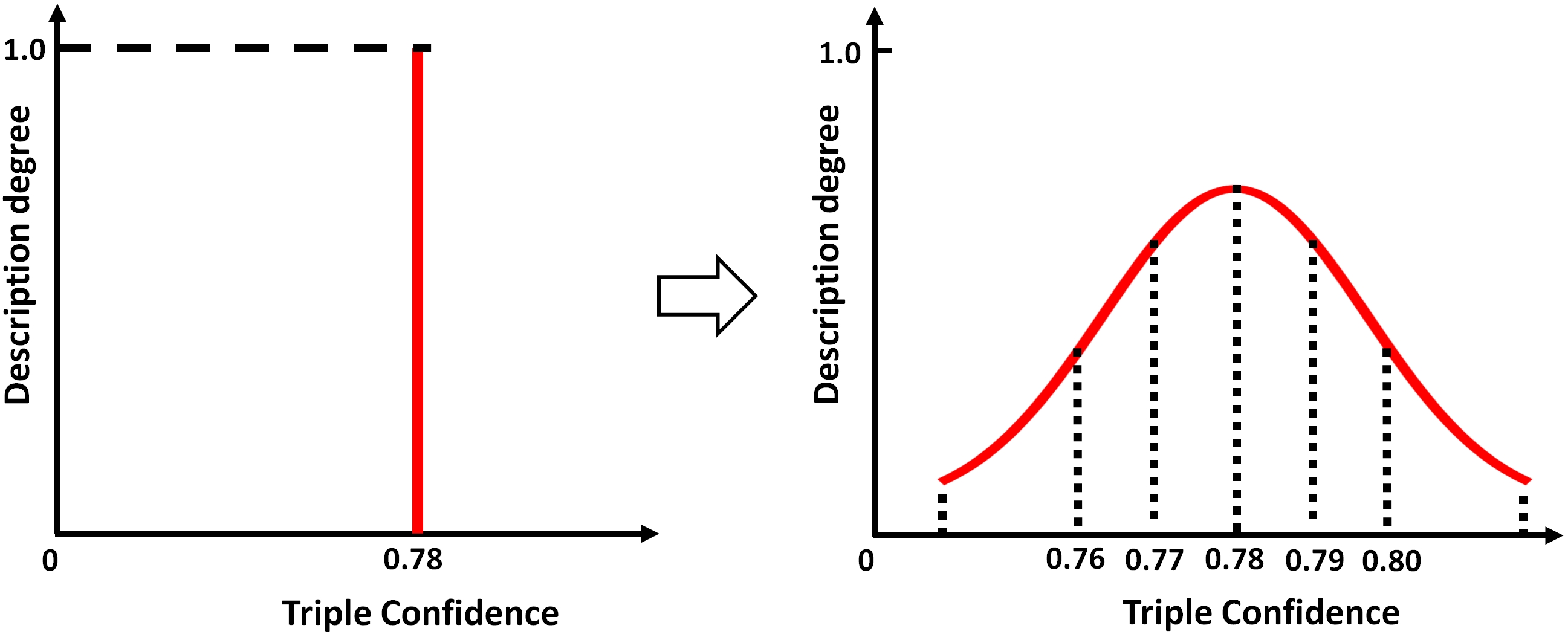}
      %  \vspace{-1mm}
    \caption{}
    \label{fig:exp_conf_1}
  \end{subfigure}
    \vspace{-2.5mm}
  \caption{(a) The histogram of triple confidence in NELL; (b) An example of a triple confidence equal to $0.78$, which is transformed to a confidence distribution.}  % 添加整体标题
  \vspace{-5mm}
  \label{fig:exp_conf}
\end{figure}

Most KGs suffer from incompleteness~\cite{DBLP:journals/dint/00800L24} since new knowledge is always emerging over time, and so are UKGs. Thus, various uncertain knowledge graph (UKG) embedding methods~\cite{chen2019embedding,kertkeidkachorn2020gtranse,chen2021passleaf,chen2021probabilistic,yang2022approximate,tseng2023upgat,wang2024unkr} are proposed to perform link prediction and confidence prediction for UKG completion. UKG embedding learns the representations of entities and relations in a low-dimensional space where graph structures and triple confidences are preserved. During the learning process, the above methods neglect the fact that the distributions of triple confidences are extremely imbalanced in most UKGs, i.e., only high-confidence triples are reserved. For example, as shown in Figure~\ref{fig:exp_conf}(a), NELL only contains the triples with confidences larger than $0.9$, and this is because there is no need to store low-confidence triples which are probably erroneous. Learning on such imbalanced data will cause that the embedding-based models cannot fit on relatively lower-confidence samples, which lowers the quality of the generated embeddings for UKG completion. In this paper, we study \textbf{how to reinforce the learning process under the imbalanced confidence distribution to obtain high-quality embeddings for UKG completion}. This problem is non-trivial, and we try to solve it by the reinforcement strategies with labeled data (i.e., existing triples with confidences) and unlabeled data (i.e., sampled unseen triples without confidences), which poses two challenges as follows:
\begin{itemize}[leftmargin=*]
    \item \textbf{Challenge 1: Reinforcement with Labeled Data.} In UKG embedding learning, labeled data are triples with confidences in the training data, and such confidences are imbalanced, so the challenge is how to effectively capture the supervision signals of unseen confidences or the confidences with a small number from labeled data to reinforce the learning process.
    \item \textbf{Challenge 2: Reinforcement with Unlabeled Data.} Since the training data is short of low-confidence triples, we aim to apply negative sampling to add unseen triples which are often false and probably low-confidence to the training data. Thus, the challenge is how to generate reliable confidences for the unlabeled data (i.e., unseen triples) to reinforce the learning process.
\end{itemize}

To solve both challenges, we propose a new \textbf{s}emi-\textbf{s}upervised \textbf{C}onfidence \textbf{D}istribution \textbf{L}earning (\textbf{ssCDL}) method for UKG completion. In ssCDL, each triple confidence is transformed into a confidence distribution. The triple confidence in UKG is a relatively fuzzy concept, e.g., there is no obvious distinction that the confidence of a triple is 0.77 or 0.78 or 0.79. This inspires us that the triples with neighboring confidences can be utilized while learning features for a particular confidence, which is similar to the usage of label distribution in facial age estimation~\cite{geng2013facial}. As shown in Figure~\ref{fig:exp_conf}(b), after transforming the confidence 0.78 into a confidence distribution for a triple in the labeled data, the supervision signals of more confidences (e.g., 0.76, 0.77, 0.79, and etc.) can be introduced into the learning process, even if such confidences are few or unseen, which \textbf{can help solve challenge 1}.

% \caption{(a) An example of a triple confidence which equals to $0.78$. (b) An example of a confidence distribution transformed from the triple confidence $0.78$.}
% \vspace{-3mm}
% \label{fig:exp_conf}
% \end{figure}

ssCDL has two components: {C}onfidence {D}istribution {L}earning based {R}elational {L}earner  ({CDL-RL}) and {P}seudo {C}onfidence {D}istribution {G}enerator ({PCDG}). CDL-RL iteratively learns UKG embeddings with labeled data and pseudo labeled data (i.e., negative sampled unseen triples with pseudo confidence labels) generated by PCDG for UKG completion. PCDG selects high-quality pseudo confidence labels for unlabeled data, and it is meta optimized by CDL-RL with labeled data. The whole process is actually meta self-training, which can alleviate the problem of gradual drifts~\cite{liu2019self} and introduce more reliable confidence labels for unlabeled data, thereby \textbf{overcoming the challenge 2}. Experiments on real-world datasets show the effectiveness and superiority of ssCDL compared with the state-of-the-art baselines in both UKG completion tasks of confidence prediction and link prediction. 

\spara{Contributions.} The main contributions of this paper are summarized as follows:
\begin{itemize}[leftmargin=*]
    \item We propose a new semi-supervised method ssCDL, which applies meta self-training to generate reliable confidences for unlabeled data in UKG embedding learning. This fully exploits unlabeled data to augment the training data so as to resolve the problem of imbalanced confidence distribution.
    \item We design a new confidence distribution learning strategy in UKG embedding learning, which transforms triple confidences into confidence distributions and this benefits to capture the supervision information of few or unseen confidences in the labeled data. 
    \item We conduct comprehensive experiments on UKG datasets, which not only shows that ssCDL outperforms baselines in different evaluation metrics for different tasks, but also verifies the effectiveness of confidence distribution learning and meta self-training for UKG completion.
\end{itemize}

\section{Related Work}
% \subsection{UKG Completion}
In this section, we review the existing studies on UKG completion, which refers to confidence prediction and link prediction. Relational learning is widely used to acquire UKG embeddings for UKG completion, and the core idea is to embed entities and relations with the structure and confidence information in the UKG, which is called normal relational learning~\cite{wang2024unkr}. Besides, few-shot relational learning further consider the long-tail distribution of relations in modeling real-world UKGs.

\spara{Normal Relational Learning for UKG}. UKGE~\cite{chen2019embedding} is a classic method in this field, which uses the scoring function of DistMult~\cite{yang2015embedding} to model triple confidences and solves the false negative problem with probabilistic soft logic. PASSLEAF~\cite{chen2021passleaf} extends UKGE for other types of scoring functions, and also uses semi-supervised learning to alleviate the false negative problem. BEUrRE~\cite{chen2021probabilistic} applies box embedding for UKG embedding learning, in which entities are represented as boxes, relationships are modeled as affine transformations of head and tail entity boxes, and triple confidences are modeled by the intersection between transformed boxes. UPGAT~\cite{tseng2023upgat} incorporates subgraph features and generalizes graph attention network for UKG completion. UKGsE~\cite{yang2022approximate} treats each triple as a short sentence and learns the confidence using LSTM. GTransE~\cite{kertkeidkachorn2020gtranse} and FocusE~\cite{DBLP:conf/ijcai/PaiC21} associate triple confidences with margin operations in the loss functions, which only solve the task of link prediction for UKGs. UKRM~\cite{chen2024uncertain} tries to mine rules using transformer to link prediction and leverage pre-trained language model to compute triple confidences. 

\spara{Few-Shot Relational Learning for UKG}. Recently, few-shot UKG completion has attracted much attention, e.g., GMUC~\cite{zhang2021gaussian} and GMUC+~\cite{wang2022incorporating} apply metric learning to learn UKG embeddings and achieve good performance, but the few-shot problem is not the focus of this paper. unKR~\cite{wang2024unkr} is a UKG embedding learning and completion tool which re-implements both works of few-shot relational learning and normal relational learning for UKG.

There also exist some works (e.g.,~\cite{fei2024soft}) regarding reasoning for query-answering on UKG, but the focus is not UKG completion, so we will not compare our method with such works. Existing relational learning methods for UKG completion neglect the fact that the distribution of triple confidences is extremely imbalanced in most UKGs, which causes the performance of UKG completion is still unsatisfactory. Our proposed method ssCDL aims to solve this problem by reinforcing the UKG embedding learning process using both labeled data and unlabeled data.

\section{Preliminaries}
% In this section, we provide detailed descriptions of the concepts of uncertainty knowledge graph completion and confidence distribution learning as follows.

%In this section, we formally define uncertain knowledge graph and uncertain knowledge graph completion, and we also introduce the concept of confidence distribution learning.

% 给出定义，加粗，描述+例子，例子文字描述
% definition，知识图谱补全写到一个定义下
% 例子斜体，强调conceptnet
\subsection{Problem Definition}
\begin{myDef}
	\label{UKG}
\spara{Uncertain Knowledge Graph}.
An uncertain knowledge graph is a repository of factual knowledge denoted as a set of quadruples $\mathcal{G} = \{(h,r,t,s)|h,t\in\mathcal{E},r\in\mathcal{R},s\in[0,1]\}$, where $\mathcal{E}$ and $\mathcal{R}$ are respectively the sets of entities and relations, and $s$ is the confidence score measuring the triple uncertainty, which represents the likelihood of the triple $(h, r, t)$ being true. 
\end{myDef}

%For example, (\textit{adults}, \textit{capableof}, \textit{drink\_beer}, \textit{0.89}) in ConceptNet means the confidence of the triple (\textit{adults}, \textit{capableof}, \textit{drink\_beer}) to be true is 0.89.
% The examples of a quadruple in UKG are list as below:
% \begin{example} \rm{Examples of Uncertain Quadruples in UKG}

% \rm{(condition, relatedto, conditional, 0.709)}

% (brunch, isa, meal, 0.893)

% (korean, languageofcountry, korea, 0.719)

% \end{example}

% \begin{table}[h!]
% \caption{Examples of Uncertain Quadruples in UKG}
% \label{example_ukg}
% \begin{tabular}{c}
% \hline
% (condition, relatedto, conditional, 0.709)  \\
% (brunch, isa, meal, 0.893) \\
% (korean, languageofcountry, korea, 0.719)\\
% \hline
% \end{tabular}
% \end{table}

\begin{myDef}
\spara{Uncertain Knowledge Graph Completion}.
Uncertain knowledge graph completion has two sub-tasks, which are confidence prediction and link prediction. Given a query $(h, r, t, ?)$ where $(h,r,t)$ is a factual triple, confidence prediction is to estimate the missing triple confidence. Given another query $(h^{'}, r^{'}, ?)$ where $h^{'}$ is a head entity and $r^{'}$ is a relation, link prediction is to predict the missing tail entity.
% TODO mydef的斜体问题
\end{myDef}

%Given a quadruple $(\textit{adults}, \textit{capableof}, \textit{drink\_beer}, \textit{0.89})$ in the UKG ConceptNet~\cite{speer2017conceptnet}, the probability of $(\textit{adults}, \textit{capableof}, \textit{drink\_beer})$ being true is $0.89$. If given the query $(\textit{adults}, \textit{capableof}, \textit{drink\_beer}, ?)$, confidence prediction aims to estimate the missing confidence as or close to $0.89$. If given the query $(\textit{adults}, \textit{capableof}, ?)$, link prediction is to predict the most likely missing tail entity $\textit{drink\_beer}$ or other possible entities.
% \begin{myDef}
% 	\label{Link Prediction}
% Given triple $(h,r,t)$, the problem of confidence prediction is to predict the confidence score of the triple based on a training set. This problem can be formally represented as $(h, r, t, ?)$. 
% \end{myDef}
% \begin{myDef}
% \label{Link Prediction}
% Given a head entity and relation pair $(h, r)$, the problem of link prediction is to predict the tail entity of the pair based on a training set. This problem can be formally represented as $(h, r, ?)$. %具体的表示形式存疑
% \end{myDef}

\subsection{Confidence Distribution Learning} \label{CDL}
Confidence distribution learning (CDL) aims to learn a model which can accurately estimate the confidence distribution of each given triple. CDL is a variant of label distribution learning~\cite{geng2016label} (LDL) applied in UKG completion. LDL is a machine learning paradigm that not only predicts the labels relevant to instances, but also quantifies the degree of relevance of each label. Before CDL, triple confidences are transformed into confidence distributions to capture the supervision information of few or unseen confidences in the labeled data.

Confidence distribution is defined as a discrete distribution in this paper, and this is also the setting of LDL~\cite{geng2016label}. In this way, since the confidence interval is $[0,1]$, we directly set the confidence labels at a granularity of $\frac{1}{n}$, and the ordered confidence label set is $\left\{0, \frac{1}{n}, \frac{2}{n}, \dots, \frac{n-1}{n}, 1 \right\}$. For a given quadruple $(h, r, t, s)$ in a UKG, we define the confidence distribution of $(h, r, t)$ as $\bm{s}=\langle s^i\rangle \in \mathbb{R}^{n+1}$, where $\sum_{i=0}^ns^i=1$ and $s^i\in[0,1]$ is a description degree that the confidence label $\frac{i}{n}$ describe the triple $(h, r, t)$. In this paper, the confidence distribution $\bm{s}$ is generated by a Gaussian distribution $\mathcal{N}(s,\sigma^2)$, where $\sigma$ is the standard deviation, and the confidence $s$ is the mean, which causes that $s$ has the highest description degree. Thus, in CDL, each piece of knowledge can be represented as a quadruple $(h,r,t, \bm{s})$, where $\bm{s} \sim \mathcal{N}(s,\sigma^2)$ is the confidence distribution. Here, we empirically set $n$ as $100$, i.e., we have $101$ confidence labels in total.

\section{Methodology}
%In this section, we introduce ssCDL, the semi-supervised confidence distribution learning method for UKG completion. 

\subsection{Overview}
% Figure \ref{fig:wide} provides the overview of ssCDL, which consists of two key components: \textbf{(1) CDL-based Relational Learner~(CDL-RL)}, which utilize confidence distribution learning to learn the embedding of head entities, relations, and tail entities, and apply it to get confidence and ranking score, so as to solve the knowledge graph completion problems of confidence prediction and link prediction. \textbf{(2) Pseudo Confidence Distribution Generator~(PCDG)}, which is based on meta self-training, generating confidence distributions for unlabeled data.

% Figure \ref{fig:wide}(a) provides the overview of ssCDL, which consists of two key components: CDL-based Relational Learner~(CDL-RL) and Pseudo Confidence Distribution Generator~(PCDG). CDL-RL and PCDG have the same structure (as shown in figure \ref{fig:wide}(b)), but their functions and training objectives are different. CDL-RL aims to learn the embedding of entites and relations with labeled data and pseudo labeled data generated by PCDG. PCDG labels and high-quality pseudo confidences for unlabeled data, and it is meta optimized by CDL-RL with labeled data. 

\begin{figure*}[htbp]
    \centering
    \includegraphics[width=\textwidth]{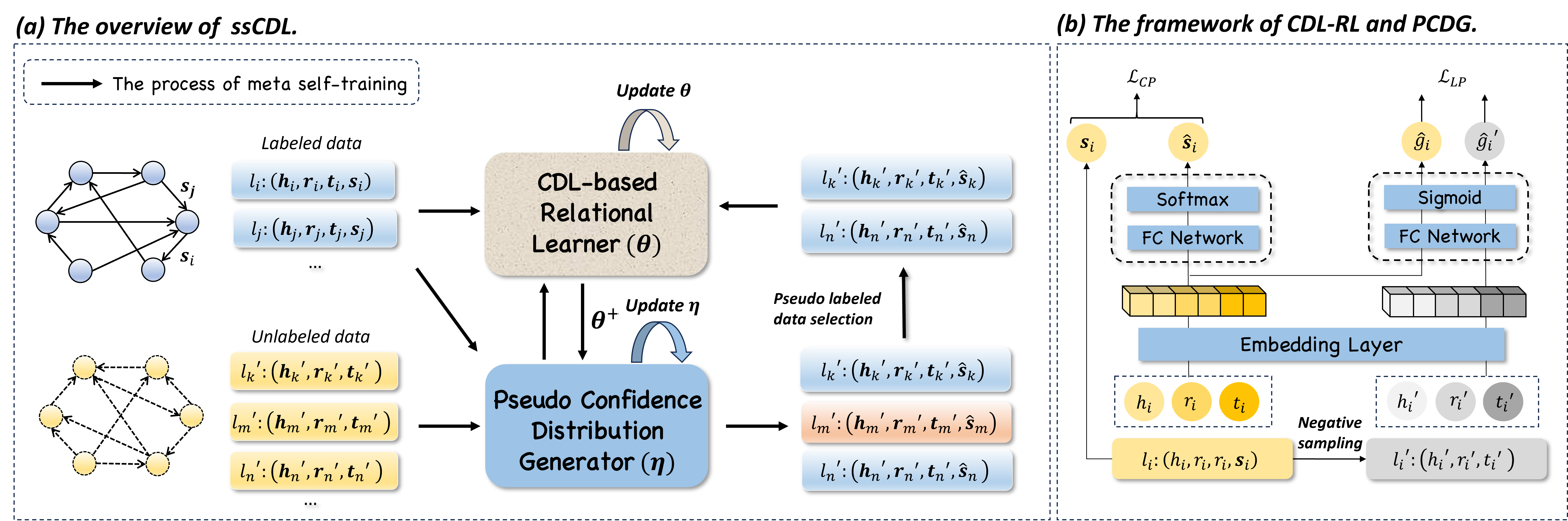} % 将example-image替换为你的图片文件名
    \caption{(a) The overview of ssCDL; (b) The framework of CDL-RL and PCDG.}
    \vspace{-5mm}
    \label{fig:wide}
\end{figure*}

Figure~\ref{fig:wide}(a) provides the overview of our semi-supervised confidence distribution learning method ssCDL, which consists of two key components: CDL-based relational learner (CDL-RL) and pseudo confidence distribution generator (PCDG). At first, we apply the strategy mentioned in Section~\ref{CDL} to transform all triple confidences in the labeled data into confidence distributions. We utilize CDL-RL to learn UKG embeddings with labeled data and pseudo labeled data generated by PCDG. PCDG generates high-quality pseudo confidences labels for unlabeled data, and CDL-RL iteratively exploits these pseudo labeled data and further improves its performance. CDL-RL and PCDG have the same structure (Figure~\ref{fig:wide}(b)), but their training processes are different. CDL-RL is optimized by minimizing the losses on confidence prediction and link prediction with labeled data and pseudo labeled data, while PCDG evaluates the performance of CDL-RL after exploiting pseudo labeled data generated by PCDG and takes it as the meta learning objective. PCDG is meta optimized by CDL-RL with labeled data. ssCDL is learned by meta self-training with iteratively training CDL-RL and PCDG.

%When one of the components is being trained, the parameters of the other component will be frozen. By iteratively training CDL-RL and PCDG, ssCDL achieves the process of meta self-training.

% Confidence distribution generator uses the loss of parameters of relational learner after one gradient descent as the optimization objective, and continuously generating pseudo labeled data during the training process. 

% The meta optimization objective of PCDG is to generate higher quality pseudo labeled data.
% The structure of confidence distribution generator and relational learner is the same, but they undertake different functions and optimization objectives. The training process of our ssCDL model is a cyclic iterative process of training CDL-RL and PCDG.
\vspace{-3.5mm}

% \subsection{UKG Embedding Learning}
\subsection{CDL-based Relational Learner}
% UKG Embedding Learner mainly consists of two modules: Confidence Distribution Predictor~(CDP) and Rank Score Predictor~(RSP). CDP can predict the confidence score and confidence distribution, enabling the UKG Embedding Learner to deal with confidence prediction. While RSP is used to predict a rank score for samples, enabling them to participate in ranking calculations in link prediction.

% To learn the embedding of entites and relations in quadruple, we utilize a UKG embedding learner. As shown in figure~\ref{}, it mainly consists of two modules, which enables the embedding learner can be optimized on both the task of confidence prediction and link prediction.
The main purpose of CDL-RL is to learn the embeddings of entities and relations. Given the $i$-th quadruple $(h_i, r_i, t_i, s_i)$ in the labeled data, it will be transformed into $l_i=(h_i, r_i, t_i, \bm{s}_i)$ as the input of CDL-RL after mapping the confidence $s_i$ to the confidence distribution $\bm{s}_i$ using a Gaussian distribution. As shown in Figure~\ref{fig:wide}(b), CDL-RL is trained by minimizing the losses on confidence prediction and link prediction, i.e., $\mathcal{L}_{CP}$ and $\mathcal{L}_{LP}$, and we explain the details as follows.
% We explain in detail how ssCDL is optimized to achieve these three objectives as follows.

% We design three learning objectives of CDL-RL: 1) minimize the difference between the predicted confidence distribution and the ground confidence distribution, 2) minimize the difference between the predicted confidence and the ground confidence, and 3) accurately predict the tail entity when given a head entity and a relation. The first two objectives are used for confidence prediction, and the last objective is for link prediction. We explain in detail how ssCDL is optimized to achieve these three objectives as follows
% The learning objective of CDL-RL is to optimize the embedding of UKG, thereby achieving accurate confidence prediction and link prediction.

To compute the loss $\mathcal{L}_{CP}$ on confidence prediction, we first concatenate the embeddings of $h_i$, $r_i$, and $t_i$ and feed it into a two-layer fully connected network (FCN), which produces an $(n+1)$-dimensional vector. We then apply the Softmax function as the activation function, and the predicted confidence distribution $\hat{\bm{s}}_i$ of the triple $(h_i,r_i,t_i)$ can be computed as:
\begin{equation} \label{predicted confidence distribution}
\hat{\bm{s}}_i=Softmax(FCN_1(\bm{h_i}||\bm{r_i}||\bm{t_i}))
\end{equation}
where $||$ represents the concatenation between embeddings, and $FCN_1$ is a function that the FCN transforms the concatenated embedding into an $(n+1)$-dimensional vector. Thus, the Kullback-Leibler (KL) divergence can be utilized to measure the similarity between the predicted confidence distribution and the ground confidence distribution. Besides, we apply the loss function of the mean squared error (MSE) between the expectation $\mathbb{E}[\hat{\bm{s}}_i]$ of predicted confidence distribution and the ground confidence $s_i$. Based on these, we formulate the learning objective of CDL-RL on confidence prediction as minimizing the KL divergence and MSE together, and define $\mathcal{L}_{CP}$ as follows:
% As shown in Figure~\ref{fig:wide}(b), given a triple $(h,r,t)$, the embeddings of $h$, $r$, and $t$ are concatenated and fed into a two-layer fully-connected network (FCN), which outputs a $(n+1)$-dimensional vector. We apply the Softmax function as the activation function. The predicted confidence distribution $\hat{\bm{s}}$ of the triple $(h,r,t)$ can be computed as $    \hat{\bm{s}}=Softmax(FCN_1(\bm{h}||\bm{r}||\bm{t}))$,
% where $||$ represents the concatenation between embeddings, and $FCN_1$ is a function that the fully-connected network transforms the concatenation of embeddings into a $(n+1)$-dimensional vector.
%
% , and make it as one of the optimization objectives of ssCDL. 
% To further improve the accuracy of confidence prediction
%Given the $i$-th quadruple $l_i$ in the training data, we denote  as the expectation of the predicted confidence distribution $\hat{\bm{s}}_i$ and take it as the predicted confidence of $l_i$.  the loss function on confidence prediction $\mathcal{L}_{CP}$ is defined as:
\begin{equation}   \mathcal{L}_{CP}=\underbrace{\sum_{\mathcal{D}}\sum_{a=0}^{n} {s_i}^a\mathrm{ln} \frac{{s_i}^a}{\hat{{s}}_i^a}}_{\text{KL divergence}} + \beta\underbrace{\sum_{\mathcal{D}} (\mathbb{E}[\hat{\bm{s}}_i]-s_i)^2}_{\text{MSE}}
\end{equation}
where ${s_i}^a$ and ${\hat{s}_i}^a$ are the $a$-th elements in ${{{\bm{s}_i}}}$ and ${\hat{{\bm{s}}}_i}$ respectively, $\mathcal{D}$ denotes the training set, and $\beta \in [0,1]$ is a hyper-parameter that controls the influence of the MSE.

To compute the loss $\mathcal{L}_{LP}$ on link prediction, we first feed the concatenated embedding of $h_i$, $r_i$, and $t_i$ into another two-layer FCN to compute the rank score of the triple $(h_i,r_i,t_i)$. This FCN outputs a single scalar normalized by the Sigmoid function as the rank score, which is computed as:
\begin{equation} \label{predicted rank score}
\hat{g}_i=Sigmoid(FCN_2(\bm{h_i}||\bm{r_i}||\bm{t_i}))
\end{equation}

% 2025.10.16之前的表述
% Here, $FCN_2$ is a function that the FCN transforms the concatenated embedding into a scalar. We then generate multiple negative samples by replacing either the head or tail entity of each positive sample with a randomly chosen entity. The number of negative samples for each positive sample is empirically set as 50. We specifically design a margin-based ranking loss function to optimize CDL-RL on link prediction, and define $\mathcal{L}_{LP}$ as follows:
% \begin{equation} \label{rankloss}
%     \mathcal{L}_{LP}=\sum_{\mathcal{D}} \sum_{\mathcal{D}^{\prime}} s_{i}[\gamma+\hat{g}_{i}^{\prime}-\hat{g}_{i}]_{+}
% \end{equation}
% where $\hat{g}_{i}^{\prime}$ is the rank score of negative sample, $\mathcal{D}^{'}$ is the set of negative samples, $[x]_{+} = \max[0,x]$ denotes the positive part of $x$, and $\gamma>0$ is a margin hyper-parameter.

Here, $FCN_2$ is a function that the FCN transforms the concatenated embedding into a scalar.  We specifically design a margin-based ranking loss function to optimize CDL-RL on link prediction, and define $\mathcal{L}_{LP}$ as follows:
\begin{equation} \label{rankloss}
    \mathcal{L}_{LP}=\sum_{\mathcal{D}} \sum_{\mathcal{D}^{\prime}} s_{i}[\gamma+\hat{g}_{i}^{\prime}-\hat{g}_{i}]_{+}
\end{equation}
% where $\hat{g}_{i}^{\prime}$ is the rank score of negative sample, $\mathcal{D}^{'}$ is the set of negative samples, $[x]_{+} = \max[0,x]$ denotes the positive part of $x$, and $\gamma>0$ is a margin hyper-parameter.
where $\hat{g}_{i}^{\prime}$ is the rank score of a negative sample which is generated by replacing either the head or tail entity of the given positive sample with a randomly chosen entity, the number of negative samples for each positive sample is empirically set as 50, $\mathcal{D}^{'}$ is the set of all negative samples, $[x]_{+} = \max[0,x]$ denotes the positive part of $x$, and $\gamma>0$ is a margin hyper-parameter.

To balance $\mathcal{L}_{CP}$ and $\mathcal{L}_{LP}$ in the training process, we use the idea of uncertainty weights~\cite{kendall2018multi} to dynamically adjust the proportion of loss for each task (we only have two tasks: confidence prediction and link prediction) during training. The final loss function of optimizing CDL-RL is defined as:
\begin{equation} \label{completeloss}
    \mathcal{L}(\mathcal{D}, \theta)=\frac{1}{2\lambda_{CP}^2}\mathcal{L}_{CP} + \frac{\phi}{2\lambda_{LP}^2}\mathcal{L}_{LP} + \log(\lambda_{CP} \cdot \lambda_{LP})
\end{equation}where $\lambda_{CP}$ and $\lambda_{LP}$ are observation noise parameters of confidence prediction and link prediction respectively, $\phi \in [0, 1]$ (empirically set as 0.1) is a weight limiting the influence of a relatively larger number of negative samples, which may cause $\mathcal{L}_{LP}$ becoming too large, and $\theta$ represents the parameters of CDL-RL.
% 2025.10.16 之前的表述
% $w_l \in [0,1]$ is the weight to limit the influence of the negative samples to the final loss function, and it is empirically set as 0.1, and $\theta$ represents the parameters of CDL-RL.

%Assuming that the output of the $k$-th task follows a Gaussian distribution, the loss of the $k$-th task can be adjusted as  $\mathcal{L}_k=\frac{1}{2\lambda_k^2}\mathcal{L}_k+\log\lambda_k$, where $\lambda_k$ is a learnable observation noise parameter of the $k$-th task, and $\mathcal{L}_k$ is the loss function of the $k$-th task. Finally, the loss function of CDL-based relational learner is defined as follows:
%\begin{equation} \label{completeloss}
%    \mathcal{L}(\mathcal{D}, \theta)=\frac{1}{2\lambda_1^2}\mathcal{L}_{CP} + \frac{1}{2\lambda_2^2}\mathcal{L}_{LP} + \log(\lambda_1 \cdot \lambda_2)
%\end{equation} where $\lambda_1$ and $\lambda_2$ are respectively observation noise parameters of confidence prediction and link prediction, and $\theta$ represents the parameters of CDL-based relational learner.

Besides the labeled data used for training CDL-RL, we apply PCDG (details will be given in Section~\ref{sec:pcdg}) to generate pseudo labeled data, which are also utilized to reinforce the training of CDL-RL. Since most pseudo labeled data are the triples with low confidences, which has almost no impact on minimizing $\mathcal{L}_{LP}$, we only use such pseudo labeled data to minimize $\mathcal{L}_{CP}$ to avoid ineffective computations. Thus, we re-define $\mathcal{L}_{CP}$ as follows:
% \begin{equation}\label{semiCDL}
% \mathcal{L}_{CP}=\sum_{\mathcal{D}_l}\sum_{a=0}^{n} {s_i}^a\mathrm{ln} \frac{{s_i}^a}{\hat{{s}}_i^a} + w_p\sum_{\mathcal{D}_p}\sum_{b=0}^{n} {s_j}^b\mathrm{ln} \frac{{s_j}^b}{\hat{{s}}_j^b} + \sum_{\mathcal{D}_l} (\mathbb{E}[\hat{\bm{s}}_i]-s_i)^2 + w_p\sum_{\mathcal{D}_p} (\mathbb{E}[\hat{\bm{s}}_j]-s_j)^2
% \end{equation}
\begin{equation}\label{semiCDL}
\mathcal{L}_{CP}=\sum_{\mathcal{D}}\sum_{a=0}^{n} {s_i}^a\mathrm{ln} \frac{{s_i}^a}{\hat{{s}}_i^a} + \beta \sum_{\mathcal{D}} (\mathbb{E}[\hat{\bm{s}}_i]-s_i)^2 + w_p(\sum_{\mathcal{D}_p}\sum_{b=0}^{n} {s_j}^b\mathrm{ln} \frac{{s_j}^b}{\hat{{s}}_j^b} + \beta \sum_{\mathcal{D}_p} (\mathbb{E}[\hat{\bm{s}}_j]-s_j)^2)
\end{equation}

% \begin{equation}\label{semiMSE}
% \mathcal{L}_{MSE}=\sum_{\mathcal{D}_l} (\mathbb{E}[\hat{\bm{s}}_i]-s_i)^2 + w_p\sum_{\mathcal{D}_p} (\mathbb{E}[\hat{\bm{s}}_j]-s_j)^2
% \end{equation}
where $\mathcal{D}_p$ denotes the set of pseudo labeled data for training CDL-RL, ${{{\bm{s}_j}}}$ and ${{{\hat{\bm{s}}_j}}}$ respectively represent the pseudo confidence distribution and the predicted confidence distribution of the $j$-th pseudo labeled quadruple $l_j$ generated by PCDG, and $w_p$ is the weight of pseudo labeled data.

% Therefore, the loss function of CDL-based relational learner with unlabeled data added can be written as:
% \begin{equation}
% \mathcal{L}(\theta)=w_{1}\mathcal{L}_{CDL}+w_{2}\mathcal{L}_{MSE}+w_{3}\mathcal{L}_{RANK}
% \end{equation}
% 考虑在这里直接表达完整的损失函数，然后说在下一节会讲如何生成unlabeled data。

\subsection{Pseudo Confidence Distribution Generator}\label{sec:pcdg}
% 四元组只在定义里出现，使用data

% 技术
% 基于未标注数据的增强，将低置信度的样本加入训练
% 为了能够有效利用未标记数据以补充样本中低置信度四元组的缺乏，我们使用了一种Meta Self-Training的技术进行数据增强，这是一种基于元学习的半监督方法。具体来说，我们通过生成负样本的方式生成unlabeled samples，然后使用Confidence Distribution Generation Model(CDG)为unlabeled data生成伪标签，从而增强训练数据并重新平衡三元组置信度的分布。
Since we aim to make full use of unlabeled data to improve the quality of UKG embedding learning in CDL-RL, we propose PCDG, which is used for generating high-quality pseudo confidence distributions for unlabeled data. PCDG is based on the idea of meta-learning, which solves the gradual shift~\cite{liu2019self} of traditional self-training. At first, we perform negative sampling to get unlabeled data (usually low-confidence triples) and feed them to PCDG. Given a positive sample in the labeled data, we generate one negative sample by replacing either the head or tail entity with a randomly chosen entity. Then, PCDG uses the most updated CDL-RL as the meta learning objective to evaluate the quality of pseudo confidence distributions by checking whether such data generated by PCDG can effectively improve CDL-RL. 
% As shown in Figure \ref{fig:wide}(a), we train an independent PCDG to generate pseudo labeled data for CDL-RL, rather than simply applying self training to CDL-RL. Note that PCDG only uses labeled data for training. Such a design effectively helps ssCDL avoid the problem of error accumulation.

% This method of training an independent module to generate pseudo labeled data, compared to traditional self-training, avoids the issue of noise accumulation caused by the noise in the generated data, which can might lead the predicted confidence distribution to drift gradually.

% To effectively leverage unlabeled data to supplement the imbalance in the training set, we employ meta self-training for data augmentation, which is a meta-learning-based semi-supervised method. The process of meta self training is shown in the figure \ref{fig:wide}, we generate unlabeled triples through the same way as negative sampling we've mentioned in section 4.2, and then utilize the Confidence Distribution Generation (CDG) Model to generate pseudo-labels for the unlabeled samples. In this way, the model can enhances the training data and rebalance the confidence distribution of the data.
%CDG具有和CDL-based Relational Learner相同的结构，但其作为元学习器被单独训练，用以为unlabeled tirples生成置信度分布。这种重新训练一个网络的方法相比于传统的使用self-training进行半监督学习的方法的好处是它能够避免单一模型在为自己生成伪标签数据时所产生的噪声数据，导致训练误差的逐渐累积。
% 直接叫relational learner
In Figure~\ref{fig:wide}(b), PCDG has the same structure as the CDL-RL, and we denote the parameters of the PCDG as $\eta$. In the training process, PCDG first generates pseudo confidence distributions for unlabeled data, and such pseudo labeled data contains gradient information of PCDG, which enables that it can be optimized. We denote these pseudo labeled data as $\mathcal{D}_{tmp}$. Note that although $\mathcal{D}_{tmp}$ and $\mathcal{D}_p$ are both generated by PCDG, they are different and so do their roles. $\mathcal{D}_{tmp}$ is generated during the phase of optimizing PCDG, while $\mathcal{D}_{p}$ is generated in the phase of training CDL-RL. Then, we minimize the loss on the labeled data (i.e., $\mathcal{D}$) after CDL-RL updates once (i.e., CDL-RL performs one gradient descent step on both $\mathcal{D}$ and $\mathcal{D}_{tmp}$). The loss function of PCDG can be expressed as:
% \begin{equation}
%     \arg\min_\eta\mathcal{L}_{labeled}(\theta^+)
% \end{equation}

% \begin{equation}
%     \arg\min_\eta\mathcal{L}(\mathcal{D}_l,\theta^+)
% \end{equation}
\begin{equation} \label{metaloss}
    \mathcal{L}(\eta)=\mathcal{L}(\mathcal{D},\theta^+)
\end{equation}

where $\mathcal{L}(\mathcal{D},\theta^+)$ (computed by Equation~(\ref{completeloss})) is the loss of CDL-RL on $\mathcal{D}$ with $\theta^+$. Here, $\theta^+$ represents the parameters of the CDL-RL after one gradient update as: % \begin{equation}
%     \theta^+=\theta-\alpha\nabla_\theta(\mathcal{L(\theta)})
% \end{equation}
\begin{equation}
    \theta^+=\theta-\alpha\nabla_\theta(\mathcal{L}(\mathcal{D} \cup \mathcal{D}_{tmp},\theta))
\end{equation}
% 这种通过对任务进行一次梯度更新来作为元优化目标的过程是元学习中常用的一种trick。这一过程在我们方法中的含义是:我们期望eta为unlabeled triples所生成的pseudo confidence distribution能够在theta的训练过程中起到效果。
where $\alpha$ is the learning rate, and $\mathcal{L}(\mathcal{D} \cup \mathcal{D}_{tmp},\theta)$ is the loss of CDL-RL on $\mathcal{D}$
and $\mathcal{D}_{tmp}$ with $\theta$.

% This process, which uses a single gradient update of the task as the meta-optimization objective, is a commonly used trick in meta-learning \cite{}. In the context of our approach, this process implies that we expect the pseudo confidence distribution generated by $\eta$ for the unlabeled triples to have a positive impact during the training of $\theta$.

% Finally, the loss function of PCDG can be explicitly expressed as:
% \begin{equation} \label{metaloss}
%     \mathcal{L}(\eta)=\mathcal{L}(\mathcal{D}_l,\theta^+)
% \end{equation}

%这里优化目标与损失函数充分

%以上可能需要进一步具体说明为什么要对theta的梯度做优化。

\subsection{Meta Self-training}
%现在我们已经介绍了CDL-RL和PCDG，本节我们将具体描述ssCDL整体的meta self-training流程。在介绍具体的meta self-training之前，我们需要介绍我们的伪标记筛选机制。
In this subsection, we introduce the overall meta self-training process of ssCDL, which iteratively trains CDL-RL and PCDG. In the process of selecting pseudo labeled data from PCDG and inputting them into CDL-RL, we apply a simple yet effective strategy. In the labeled data, the original confidence of each triple should take the lead in the transformed confidence distribution. Thus, in the pseudo labeled data, if the highest description degree of a pseudo confidence label is larger than a fixed threshold, the corresponding pseudo confidence distribution is treated as high-quality and the pseudo labeled data will be selected for training CDL-RL, while others will be removed.

% In the process of meta self-training, the PCDG generates pseudo labeled data for CDL-RL, and CDL-RL provides meta learning objective for PCDG. This iterative training process is our complete meta self-training.
% we first feed labeled data $\mathcal{D}_l$ into CDL-RL and utilize CDL-RL to learn UKG embeddings, CDL-RL optimize itself by minimizing losses on the tasks of confidence prediction and link prediction (Equation~\ref{completeloss}). 
% When the embedding of UKG gradually stabilizes, we start to train PCDG, which aims to generate high-quality pseudo confidences labels for unlabeled data, and it is optimized by the loss of CDL-RL after performing one gradient descent step on both the labeled data and pseudo labeled data $\mathcal{D}_{tmp}$. After training PCDG for some time, we apply the selection strategy to pseudo labeled data generated by PCDG and feed these selected pseudo labeled data $\mathcal{D}_{p}$ into CDL-RL and enhance its training process. Afterwards, these two components will continuously repeat the above training process and optimize themselves, i.e., 

Algorithm~\ref{alg:mst} gives the details on meta self-training of ssCDL. 
% 原先的描述
We define two important time points, i.e., the epoch of starting training PCDG $T_{PCDG}$ and the epoch of starting using pseudo labeled data for CDL-RL $T_{CDLRL}$. We first initialize the parameters $\theta$ and $\eta$ of CDL-RL and PCDG respectively, and take the labeled data $\mathcal{D}$ and sampled unlabeled data $\mathcal{D}_u$ as the input of ssCDL. 
% 后来修改的描述
% For each labeled data in $\mathcal{D}_l$, we generate an unlabeled data by replacing either the head or tail entity of each positive data with a randomly chosen entity, thus obtaining the set of unlabeled data $\mathcal{D}_u$. We define two important time points, i.e., the epoch to start training PCDG $T_{PCDG}$ and the epoch to start using pseudo labeled data for CDL-RL $T_{CDLRL}$.
% We first initialize the parameters $\theta$ and $\eta$ of CDL-RL and PCDG respectively, and feed $\mathcal{D}_l$ and $\mathcal{D}_u$ as the input of ssCDL. 
When the number of the current epoch $N_{cur}$ is less than $T_{PCDG}$, we only need to optimize CDL-RL with $\mathcal{D}$ (\textit{line 2-4}). This setting tries to make UKG embeddings become stable in this period, which will help us train PCDG soon. When $N_{cur}$ is larger than or equal to $T_{PCDG}$ but less than $T_{CDLRL}$, it indicates that UKG embeddings have stabilized, and we start to train PCDG. PCDG first generates $\mathcal{D}_{tmp}$ for unlabeled data, and feeds $\mathcal{D}_{tmp}$ and $\mathcal{D}$ together into the most updated CDL-RL, so we get the parameters $\theta^+$ of CDL-RL updated for one more time (\textit{line 6-7}). PCDG will utilize $\theta^+$ to update itself with $\mathcal{D}$ (\textit{line 8}). In this period, we do not directly use PCDG to generate $\mathcal{D}_p$ and feed it to CDL-RL because in the early stages of PCDG training, the quality of the generated labels cannot be guaranteed. Therefore, the training process of CDL-RL is the same as before (\textit{line 9}). If $N_{cur}$ is larger than or equal to $T_{CDLRL}$, we will exploit pseudo labeled data $\mathcal{D}_p$ to help train CDL-RL (\textit{line 11-18}). We use PCDG to generate $\mathcal{D}_p$ and apply the pseudo label selection strategy, and the selected $\mathcal{D}_p$ will be used together with $\mathcal{D}$ to optimize CDL-RL, while the optimization of PCDG is the same as before (\textit{line 12-14}). Afterwards, PCDG and CDL-RL will continuously repeat the above training process and optimize themselves until the maximum epoch is reached, i.e., PCDG generates pseudo labeled data for CDL-RL, and CDL-RL provides the meta learning objective for PCDG. This iterative training process is our complete meta self-training. During training, all parameters including embeddings, are updated using stochastic gradient descent (SGD) in each minibatch.

\begin{algorithm}
\caption{Meta Self-Training of ssCDL}
\label{alg:mst}
\DontPrintSemicolon
\SetAlgoLined % 显示行号
\SetNlSty{}{}{} % 可选：自定义行号样式
% 别忘了所有的行数都要变
\KwIn{Labeled data $\mathcal{D}$, unlabeled data $\mathcal{D}_u$, the parameters of CDL-RL $\theta$, the parameters of PCDG $\eta$, the number of current epoch $N_{cur}=1$, the number of maximum epoch $T_{max}$, the epoch of starting training PCDG $T_{PCDG}$, the epoch of starting using pseudo labeled data for CDL-RL $T_{CDLRL}$, the learning rate $\alpha$.}
\BlankLine
\While{$N_{cur}$~$\leq$~$T_{max}$}{

    % \textbf{Step 1: Optimize CDL-RL} \\
    
    \If {$N_{cur}$~$<$~$T_{PCDG}$}{
$\theta \leftarrow \theta - \alpha\nabla_{\theta} \mathcal{L}(\mathcal{D}, \theta)$; \hfill\makebox[0pt][r]{\scriptsize\ensuremath{\triangleright}~\textit{Update CDL-RL with labeled data.}}\\
    }
    \If {$T_{PCDG}$~$\leq$~$N_{cur}$~$<$~$T_{CDLRL}$}{
    % \textbf{Meta update PCDG with labeled data.}\\
    % // \textit{Meta update PCDG with labeled data.}\\
    % {\scalebox{0.8}{Meta update PCDG with labeled data}}\\

    % Generate pseudo labeled data for training PCDG: \\
    $\mathcal{D}_{tmp} = f_{\eta}(\mathcal{D}_u)$;
    \\
     $\theta^+ \leftarrow \theta - \alpha\nabla_{\theta} \mathcal{L}(\mathcal{D}\cup \mathcal{D}_{tmp},\theta)$;\\
    $\eta \leftarrow \eta - \alpha\nabla_{\eta} \mathcal{L}(\mathcal{D}, \theta^+)$; \hfill\makebox[0pt][r]{\scriptsize\ensuremath{\triangleright}~\textit{Meta update PCDG with labeled data.}}\\
    % \textbf{Update CDL-RL with labeled data.}\\

    $\theta \leftarrow \theta - \alpha\nabla_{\theta} \mathcal{L}(\mathcal{D}, \theta)$;\hfill\makebox[0pt][r]{\scriptsize\ensuremath{\triangleright}~\textit{Update CDL-RL with labeled data.}}\\
    % Freeze $\theta$\\
    % Freeze $\eta$\\
    }
    \If {$N_{cur}$~$\geq$~$T_{CDLRL}$}{
    % \textbf{Meta update PCDG with labeled data.}\\
    % // \textit{Meta update PCDG with labeled data.}\\

    % Generate pseudo labeled data for training PCDG: //
    $\mathcal{D}_{tmp} = f_{\eta}(\mathcal{D}_u)$; 
    \\
    % $\theta^+ \leftarrow \theta - \nabla_{\theta} (\mathcal{L}(\mathcal{D}_l,\theta) +  w_p\mathcal{L}(\mathcal{D}_{tmp}, \theta))$\\
    $\theta^+ \leftarrow \theta - \alpha\nabla_{\theta} \mathcal{L}(\mathcal{D}\cup \mathcal{D}_{tmp},\theta)$;\\
    $\eta \leftarrow \eta - \alpha\nabla_{\eta} \mathcal{L}(\mathcal{D}, \theta^+)$; \hfill\makebox[0pt][r]{\scriptsize\ensuremath{\triangleright}~\textit{Meta update PCDG with labeled data.}}\\
    % // \textit{Update CDL-RL with labeled data and pseudo labeled data.}\\

        % Generate pseudo labeled data for training CDL-RL:\\
    $\mathcal{D}_{p} = f_{\eta}(\mathcal{D}_u)$;\\
    $\mathcal{D}_p = \text{Select}(\mathcal{D}_p)$ ;\hfill\makebox[0pt][r]{\scriptsize\ensuremath{\triangleright}~\textit{Pseudo labeled data selection.}}\\
    $\theta \leftarrow \theta - \alpha\nabla_{\theta} \mathcal{L}(\mathcal{D}\cup\mathcal{D}_p, \theta)$;\hfill\makebox[0pt][r]{\scriptsize\ensuremath{\triangleright}~\textit{Update CDL-RL with labeled data and pseudo labeled data.}}}
    $N_{cur}++$;
    % Freeze $\theta$\\
    % Freeze $\eta$\\
    % }
    % \quad Generate pseudo labeled data: ${D}_p = f_{\eta}(D_u)$ \\
    % \quad Apply pseudo labeled data selection: ${D}_p = \text{Select}({D}_p)$ \\
    % \quad Update relational learner: $\theta \leftarrow \theta - \nabla_{\theta} (\mathcal{L}(X_l, Y_l; \theta) +  w_u\mathcal{L}(X_{u1}, \hat{Y}_{u1}^{\text{high}}; \theta))$ \\
    % \BlankLine
    % % \textbf{Step 2: PCDG} \\
    % \quad Generate pseudo-labels: $\hat{Y}_{u2} = f_{\eta}(X_{u2})$ for the the training of CDG.\\

    % \quad Compute updated relational learner:
    % $\theta^+ \leftarrow \theta - \nabla_{\theta} (\mathcal{L}(X_l, Y_l; \theta) +  w_u\mathcal{L}(X_{u2}, \hat{Y}_{u2}^{\text{high}}; \theta))$

    % \quad Meta-update pseudo-label generator: $\eta \leftarrow \eta - \alpha\nabla_{\eta} \mathcal{L}(X_l, Y_l; \theta^+)$ \\
}
\end{algorithm}
%\vspace{-3mm}

\section{Experiments}
% In this section, we first introduce our experimental procedure and
% settings, 
In this section, we present experiments to show the effectiveness and superiority of ssCDL on the UKG completion tasks of confidence prediction and link prediction. We also analyze the effects of CDL and meta self-training in ssCDL with ablation experiments, and further explore the performance of ssCDL on predicting the confidences of low-confidence triples. We also investigate the sensitivity of three key hyper-parameters of ssCDL, which is shown in Appendix~\ref{app:PS}. The source code of ssCDL is publicly available at: \url{https://github.com/seucoin/unKR/tree/main/unKR_ssCDL}.

\textbf{Datasets}. We conducted experiments on two widely used UKG datasets, i.e., NL27k extracted from NELL and CN15k extracted from ConceptNet (more details are in Table~\ref{dataset}). We followed the setting of UKGE~\cite{chen2019embedding} to partition the dataset into 85\% for training, 7\% for validation, and 8\% for testing. 
% \begin{itemize}
%     \item \textbf{CN15k} is a subgraph of the commonsense KG ConceptNet. CN15k limits the confidence of ConceptNet to the interval of [0.1, 3.0] and maps the confidence scores of these uncertain triples to [0.1, 1.0] through min-max normalization.
    
%     \item \textbf{NL27k} is extracted from NELL. Nl27k maps the confidences of NELL from the interval of [0.9, 1.0] to the interval of [0.1, 1.0].
% \end{itemize}

% \justificationTODO{let table in the text}

\spara{Baselines}. We compared ssCDL with the state-of-art embedding-based UKG completion methods, including  \textbf{UKGE}~\cite{chen2019embedding}, \textbf{PASSLEAF}~\cite{chen2021passleaf}, \textbf{UKGsE}~\cite{yang2022approximate}, \textbf{BEUrRE}~\cite{chen2021probabilistic}, and \textbf{UPGAT}~\cite{tseng2023upgat}, in confidence prediction and link prediction. More details of such baselines are given in Appendix~\ref{app:A1}.

\subsection{Experimental Setup}
\label{ex:setup}

\spara{Implementation Details}. 
% \begin{wraptable}{r}{0.5\textwidth}
%     \hspace{-2em} % 添加这一行左移一点点，具体数值可调
%     \centering
%     \captionsetup{font=small}
%     \caption{Statistics of UKG Datasets.}
%     \label{dataset}
%     \vspace{0.2em}
%     \begin{tabular}{c | c c c}
%         \toprule
%         Dataset & \#Entities & \#Relations & \#Quadruples \\
%         \midrule
%         NL27k & 27,221 & 404 & 175,412 \\
%         CN15k & 15,000 & 36 & 241,158 \\
%         \bottomrule
%     \end{tabular}
% \end{wraptable}
 ssCDL was implemented by Pytorch-lightning, and all the experiments were conducted on an RTX3090 GPU card. The optimal hyper-parameters of ssCDL on NL27k are as follows: the standard deviation of Gaussian distribution $\sigma=0.6$, the weight of pseudo labeled data $w_p=0.7$, the MSE weight $\beta=1$, the learning rate $\alpha=0.001$, the margin $\gamma=0.1$, the threshold for 
 \hspace{-3em} % 添加这一行左移一点点，具体数值可调
 \begin{wraptable}{r}{0.53\textwidth}
    \hspace{-3em} % 添加这一行左移一点点，具体数值可调
    \centering
    \captionsetup{font=small}
    %\vspace{-5mm}
    \caption{The statistics of UKG Datasets.}
    \label{dataset}
    \vspace{0.2em}
    \begin{tabular}{c | c c c}
        \toprule
        Dataset & \#Entities & \#Relations & \#Quadruples \\
        \midrule
        NL27k & 27,221 & 404 & 175,412 \\
        CN15k & 15,000 & 36 & 241,158 \\
        \bottomrule
    \end{tabular}
    %\vspace{-2mm}
\end{wraptable}
pseudo labeled data selection: 0.03, the batch size: 4096, and the embedding size: 128. The optimal hyper-parameters of ssCDL on CN15k are as follows:
the standard deviation of Gaussian distribution $\sigma=0.6$, the weight of pseudo labeled data $w_p=0.3$, the MSE weight $\beta=1$, the learning rate $\alpha=0.001$, the margin $\gamma=0.1$, the threshold for pseudo labeled data selection: 0.015, the batch size: 4096, and the embedding size: 512.
We conducted sensitivity analysis on $\sigma$, $w_p$, and the threshold for pseudo labeled data selection, and more details are given in Appendix~\ref{app:PS}. The experimental results of all baselines refer to the implementation of unKR~\cite{wang2024unkr}, and more details are given in Appendix~\ref{app:ID}.

% \spara{Evaluation Protocol:}
\spara{Evaluation Protocol}. We evaluated ssCDL and baselines on the tasks of confidence prediction and link prediction with the following evaluation metrics. For confidence prediction, we selected \textbf{Mean Squared Error (MSE)}: the mean squared error of predicted confidences and ground confidences, and \textbf{Mean Absolute Error (MAE)}: the mean absolute error of predicted confidences and ground confidences. For link prediction, we chose \textbf{Hits@1}: the proportion of ranks equal to one for all tail entities, and \textbf{Weighted Mean Reciprocal Rank (WMRR)}: the weighted average multiplicative inverse of the ranks for all tail entities.

\subsection{Confidence Prediction}
% As we've mentioned in section \ref{UKG}, confidence prediction aims to predict the confidence for triples. 
%Confidence prediction in UKGs is typically defined as the task of predicting the confidence for a triple, i.e., given a query $(h,r,t,?)$, confidence prediction aims to estimate the missing triple confidence of $(h,r,t)$.

We compared ssCDL with all baselines on CN15k and NL27k in the task of confidence prediction. As shown in Table~\ref{main_result}, ssCDL outperforms all baselines in both MSE and MAE. Compared with the best baseline on NL27k, ssCDL reduces MSE and MAE by 52.6\% and 17.6\%, respectively. Compared with the best baseline on CN15k, ssCDL reduces MSE and MAE by 63.8\% and 53.2\%, respectively. It demonstrates that ssCDL can effectively capture the semantics, structure, and confidence information in UKGs, enabling more accurate prediction of triple confidences.

We noticed that all methods' performance on NL27k are better than that on CN15k. This can be attributed to the triple confidences in ConceptNet are determined solely by its data sources and the frequency with which they are mentioned. While different data sources may be of different quality, most triples in ConceptNet are generally correct. As a result, there is no significant distinction in reliability between high-confidence and low-confidence triples in ConceptNet. This situation may cause that the performance of all methods on CN15K is not that good, so it is necessary to build better UKG completion benchmark datasets.

\begin{table*}[h!]
\centering
\setlength{\tabcolsep}{5pt}
\caption{The comparison results between ssCDL and baselines on NL27k and CN15k for confidence prediction and link prediction. The best results are indicated by bold numbers, while the runner-up results are indicated by underlined numbers (CP: confidence prediction, LP: link prediction).}
\label{main_result}
\begin{tabular}{c | c c | c c | c c | c c}
\midrule Dataset & \multicolumn{4}{c|}{NL27k} & \multicolumn{4}{c}{CN15k} \\
\midrule
Task & \multicolumn{2}{c|}{CP} & \multicolumn{2}{c|}{LP} & 
 \multicolumn{2}{c|}{CP} & \multicolumn{2}{c}{LP}\\
\midrule Metric & MSE&MAE&WMRR&Hits@1& MSE &MAE&WMRR&Hits@1 \\ 
\midrule
UKGE$_{logi}$ & 0.029 & 0.060 & 0.593 & 0.462 & 0.246 & 0.409 & 0.118 & 0.072\\
UKGE$_{rect}$ & 0.033 & 0.071 & 0.580 & 0.452& 0.202 & 0.364 & 0.127 & 0.060 \\
BEUrRE & 0.089 & 0.222 & 0.272 & 0.117& 0.117 & 0.283 & 0.138 & 0.039 \\
PASSLEAF$_{DistMult}$ & 0.023 & \underline{0.051} & 0.676 & 0.553& 0.216 & 0.379 & 0.170 & 0.078 \\
PASSLEAF$_{ComplEx}$ & 0.024 & 0.052 & 0.708 & \underline{0.586}& 0.231 & 0.400 & \underline{0.196} & \underline{0.086} \\
PASSLEAF$_{RotatE}$ & \underline{0.019} & 0.063 & \underline{0.715} & 0.580& \underline{0.094} & \underline{0.248} & 0.137 & 0.037 \\
UKGsE & 0.122 & 0.271 & 0.064 & 0.031 & 0.103 & 0.256 & 0.012 & 0.002 \\
UPGAT & 0.029 & 0.101 & 0.658 & 0.530 & 0.149 & 0.308 & 0.165 & 0.078 \\
\midrule
\textbf{ssCDL} & \textbf{0.009} & \textbf{0.042} & \textbf{0.727} & \textbf{0.636}& \textbf{0.034} & \textbf{0.116} & \textbf{0.207} & \textbf{0.133} \\
\midrule

% \hline Task & \multicolumn{4}{|c|}{CP} & \multicolumn{4}{|c|}{LP} \\
% \hline Metric & MSE&MAE&MSE&MAE& WMRR&Hit@1&WMRR&Hit@1 \\ \hline
% w/o semi & 0.012 & 0.040 & 0.037 & 0.117 & 0.724 & 0.628 & 0.201 & 0.127\\
% w/o cdl & 0.015 & 0.057 & 0.044 & 0.141& 0.586 & 0.482 & 0.149 & 0.090 \\
% \hline
% \textbf{ssCDL} & \textbf{0.012} & \textbf{0.037} & \textbf{0.037} & \textbf{0.115}& \textbf{0.731} & \textbf{0.635} & \textbf{0.207} & \textbf{0.129} \\
\end{tabular}
\end{table*}

\subsection{Link Prediction}
%Link prediction in UKGs is typically defined as the task of predicting the most likely missing tail entity for given head entity and relation, i.e., given a query $(h,r,?)$, link prediction aims to predict the missing tail entity.
% \justificationTODO{need better discussion}
Table~\ref{main_result} also presents the comparison results on link prediction, and ssCDL achieves the best Hits@1 and WMRR, which demonstrates that ssCDL is capable of predicting more accurate tail entities for different queries. Compared with confidence prediction, ssCDL does not have a quite significant improvement on link prediction compared with baselines. This is because our designed CDL and meta self-training are used to optimize confidence prediction to get better UKG embedding, and link prediction only benefits from such embeddings besides minimizing the margin-based ranking loss. This relatively implicit optimization may cause the improvement in link prediction is not as significant as that in confidence prediction. However, ssCDL still outperforms all baselines on link prediction, indicating that the training of ssCDL is an effective multi-task learning process.

\subsection{Ablation Study}\label{subsec:as}
To investigate the contributions of CDL and meta self-training in ssCDL, we conducted ablation study on both datasets, and the results are given in Table~\ref{abl}.

\spara{Ablation on Confidence Distribution Learning.} To evaluate the effectiveness of our CDL strategy, we did not use confidence distributions of triples, but only utilized their ground confidences for training (denoted as w/o cdl). This variant shows a significant performance decline on both datasets, which confirms the effectiveness of CDL. This also indicates that confidence distributions enable ssCDL to better utilize the supervision information of different confidences in the training data and reinforce the embedding learning process.

%transformed each confidence of triples into a one-hot encoding rather than a confidence distribution (i.e., w/o cdl). 

\spara{Ablation on Meta Self-Training.} To verify the effectiveness of meta self-training, we removed PCDG, and only applied labeled data to train CDL-RL (denoted as w/o mst). This variant also exhibits performance decline, demonstrating that the pseudo labeled data generated by PCDG do assist ssCDL in learning better UKG embeddings. Besides, w/o mst outperforms w/o cdl, illustrating that the supervision information of the labeled data has a more important influence on the entire learning process compared to the potential information of the unlabeled data.

\begin{table*}[h!]
\centering
\setlength{\tabcolsep}{8pt}
\caption{The ablation study of ssCDL on confidence prediction and link prediction. The best results are highlighted in bold. w/o cdl refers to removing confidence distribution learning, and w/o mst refers to removing meta self-training (CP: confidence prediction, LP: link prediction).}
\label{abl}
\begin{tabular}{c | c c | c c | c c | c c}
\midrule Dataset & \multicolumn{4}{c|}{NL27k} & \multicolumn{4}{c}{CN15k} \\
\midrule
Task & \multicolumn{2}{c|}{CP} & \multicolumn{2}{c|}{LP} & 
 \multicolumn{2}{c|}{CP} & \multicolumn{2}{c}{LP}\\
\midrule Metric & MSE&MAE&WMRR&Hits@1& MSE &MAE&WMRR&Hits@1 \\ 
\midrule
w/o cdl & 0.015 & 0.057 & 0.586 & 0.482& 0.044 & 0.141 & 0.149 & 0.090 \\
w/o mst & 0.010 & 0.045 & 0.718 & 0.619 & 0.035 & 0.118 & 0.200 & 0.128\\
\midrule
\textbf{ssCDL} & \textbf{0.009} & \textbf{0.042} & \textbf{0.727} & \textbf{0.636}& \textbf{0.034} & \textbf{0.116} & \textbf{0.207} & \textbf{0.133} \\
\midrule
% \hline Task & \multicolumn{4}{|c|}{CP} & \multicolumn{4}{|c|}{LP} \\
% \hline Metric & MSE&MAE&MSE&MAE& WMRR&Hit@1&WMRR&Hit@1 \\ \hline
% w/o semi & 0.012 & 0.040 & 0.037 & 0.117 & 0.724 & 0.628 & 0.201 & 0.127\\
% w/o cdl & 0.015 & 0.057 & 0.044 & 0.141& 0.586 & 0.482 & 0.149 & 0.090 \\
% \hline
% \textbf{ssCDL} & \textbf{0.012} & \textbf{0.037} & \textbf{0.037} & \textbf{0.115}& \textbf{0.731} & \textbf{0.635} & \textbf{0.207} & \textbf{0.129} \\
\end{tabular}
\end{table*}

\subsection{Low-Confidence Triples Analysis}\label{subsec:low}
%为了证明ssCDL能够缓解传统UKG补全方法在低置信度三元组上的类别不平衡问题，我们在UKG的低置信度三元组上也进行了实验。cn15k与nl27k中近80%的三元组置信度高于0.5，我们分别随机从cn15k与nl27k中采样50个置信度低于0.5的三元组进行评测。

% 在正文中插入环绕图
As mentioned before, ssCDL is designed to reinforce the UKG embedding learning process under the imbalanced confidence distribution. Since nearly 80\% of the triples in CN15k and NL27k have 
\begin{wrapfigure}{r}{0.55\textwidth}
  \centering
  \begin{subfigure}[b]{0.26\textwidth}
    \includegraphics[width=\textwidth]{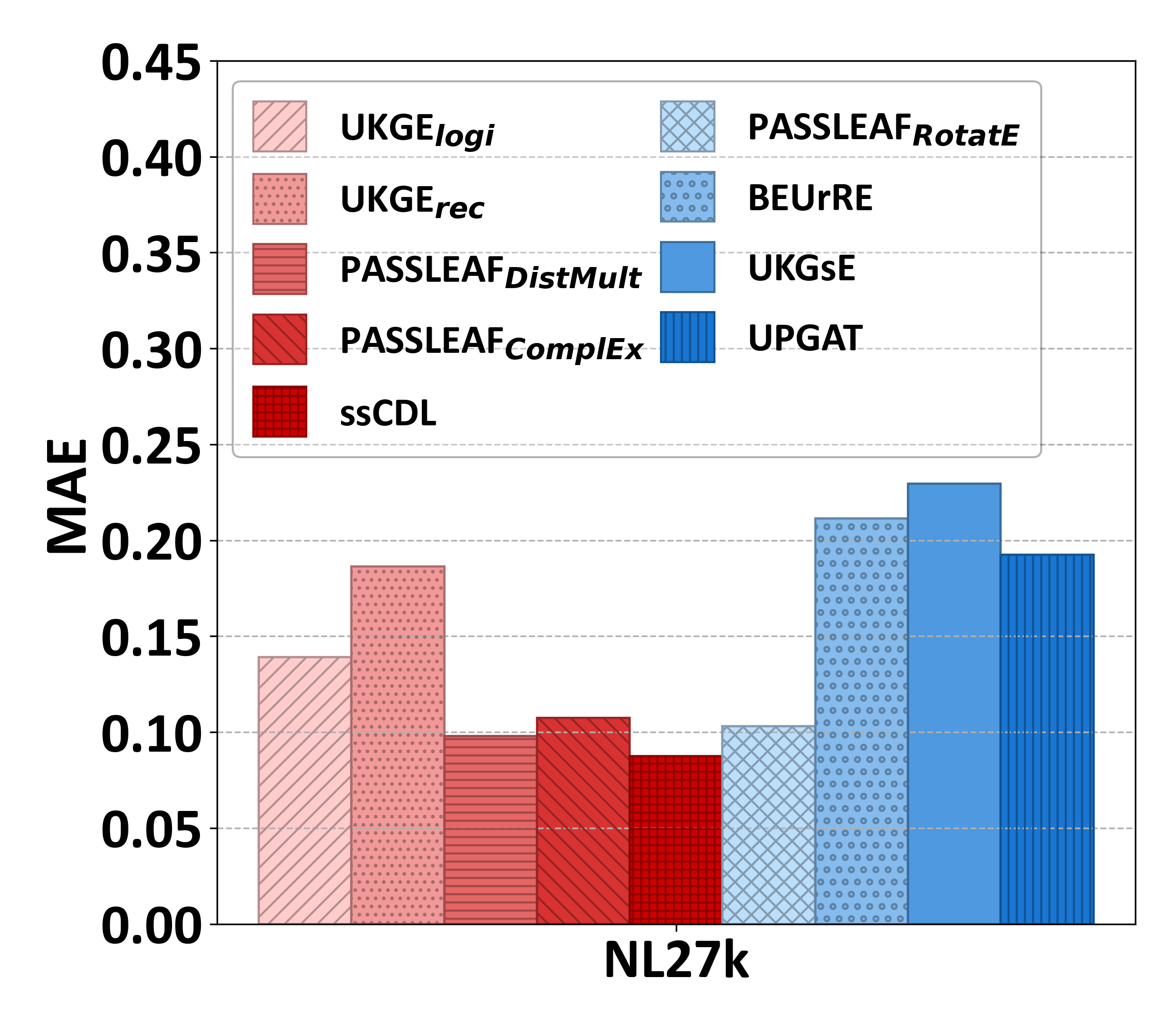}
    \caption{}
  \end{subfigure}
  \hfill
  \begin{subfigure}[b]{0.26\textwidth}
    \includegraphics[width=\textwidth]{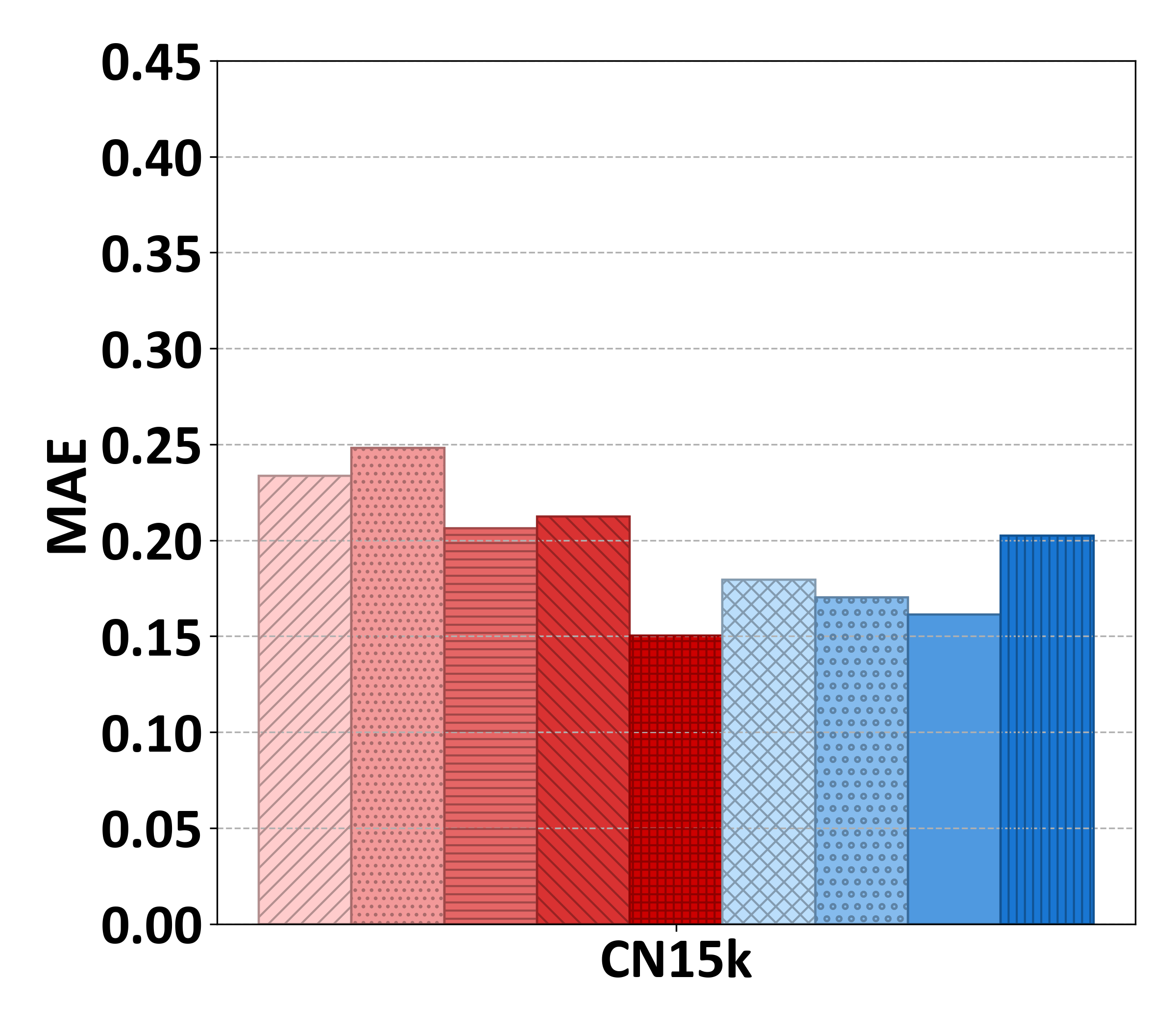}
    \caption{}
    \label{fig:exp_conf_b}
  \end{subfigure}
  \caption{The comparison results of confidence prediction on low-confidence triples in NL27k and CN15k.}
 \label{fig:lowconf}
 \end{wrapfigure}
the confidence higher than 0.5, we conducted the experiments on predicting the confidences of actual low-confidence triples to demonstrate that ssCDL can alleviate the problem of imbalanced confidence distribution in confidence prediction. Since low-confidence triples are less reliable, it is unnecessary to perform link prediction. We randomly selected the triples with the confidences lower than 0.5 from the test set of each dataset, and used MAE to evaluate the performance on confidence prediction. The comparison results, as illustrated in Figure~\ref{fig:lowconf}, reveal that most baselines are not good at handling confidence prediction on low-confidence triples, while ssCDL still demonstrates the robustness and outperforms all baselines on both datasets. This fully reflects the effectiveness of ssCDL in predicting the confidences of low-confidence triples. 

%Besides, semi-supervised learning methods including the variants of PASSLEAF and ssCDL, consistently outperform other baselines, which indicates that the generated pseudo labeled data can alleviate the problem of imbalanced confidence distribution in UKG embedding learning for confidence prediction.
%randomly sampled fifty triples with the confidences lower than 0.5 from both datasets respectively for evaluation. Since link prediction on low-confidence triples are meaningless, we only report results on confidence prediction in this subsection. The model trained on the complete training set are tested on the sampled low-confidence triples and MAE is selected to evaluate the performance on confidence prediction. The results, as illustrated in Figure \ref{fig:lowconf}, reveal that most baselines are inefficient in handling confidence prediction on low-confidence triples, while ssCDL still demonstrates robustness and outperforms all baselines. The results suggest that the strategy used by ssCDL effectively alleviates the imbalanced confidence distribution problem in UKG. Meanwhile, semi-supervised learning-based methods (including variants of PASSLEAF and ssCDL) generally outperform other types of baselines, which reflects that the pseudo labeled data generated through semi-supervised learning-based method can alleviate the problem of imbalanced confidence distribution.

\section{Conclusion}
In this paper, we propose a new semi-supervised confidence distribution learning method ssCDL for UKG completion. ssCDL is composed of CDL-based relational learner and pseudo confidence distribution generator, which are iteratively trained by meta self-training. Such a semi-supervised learning framework and the introduction of triple confidence distributions benefit to solve the problem of extremely imbalanced distributions of triple confidences in UKG embedding learning for UKG completion. Experimental results demonstrate that ssCDL has the best performance on real-world UKG datasets compared with the state-of-the-art baselines. The effectiveness of different strategies used in ssCDL has been verified in the ablation study.

As for the future work, we plan to study rule learning and reasoning on UKGs based on confidence distribution learning. We will also explore to apply large language model to UKG completion, and use UKGs for reliable retrieval-augmented generation.

\section{Acknowledgements}
This work is supported by the NSFC (Grant No. 62376058, 52378009, 62576093, U23B2057, 62176185, 62476058), the Southeast University Interdisciplinary Research Program for Young Scholars, and the Big Data Computing Center of Southeast University.

\bibliographystyle{plain}
\bibliography{nips}

\newpage
\appendix

\section{Baselines} \label{app:A1}
We compared ssCDL with the state-of-art embedding-based UKG completion methods published in recent years, including:

\begin{itemize}
    \item \textbf{UKGE}~\cite{chen2019embedding} is a classic UKG completion method, which is the first work on relational learning on UKG and it has two variants: UKGE$_{logi}$ using the logistic function as the mapping function, and UKGE$_{rect}$ which takes a bounded rectifier as the mapping function.
    \item \textbf{PASSLEAF}~\cite{chen2021passleaf} improves the generalization ability of UKGE and applies semi-supervised learning for the first time in UKG completion to solve the false negative problem. Now it is the best UKG completion method. PASSLEAF$_{DistMult}$, PASSLEAF$_{ComplEx}$, and PASSLEAF$_{RotatE}$ are three variants of PASSLEAF, which use the scoring functions of DistMult~\cite{yang2015embedding}, ComplEx~\cite{trouillon2016complex} and RotatE~\cite{sun2019rotate}, respectively.
    \item \textbf{UKGsE}~\cite{yang2022approximate} treats each knowledge fact as a short sentence, and is a typical model of UKG completion leveraging a pre-trained language model.
    \item \textbf{BEUrRE}~\cite{chen2021probabilistic} is a representative UKG completion model based on box embedding. The geometry of boxes endows the model with calibrated probabilistic semantics and facilitates the incorporation of relational constraints.
    \item \textbf{UPGAT}~\cite{tseng2023upgat} generalizes the graph attention network and uses it to capture the local structural information in UKG completion. It is a typical UKG completion method modeling structural contexts in UKG.
\end{itemize}

\section{Implementation Details. } \label{app:ID}
We used Adam optimizer~\cite{adam} for SGD training. For hyper-parameter tuning, we searched the best hyper-parameter from the following settings: the batch size $\in$ \{512, 1024, 2048, 4096\}, the embedding size $\in$ \{128, 256, 512\}, the margin $\gamma$ $\in$ \{0.01, 0.05, 0.1, 0.5, 1\}, and the MSE weight $\beta$ $\in$ \{0.6, 0.8, 1\}. We conducted detailed sensitivity analysis on the following parameters, including the standard deviation of Gaussian distribution $\sigma$ $\in$ \{0.02, 0.2, 0.6, 1, 2\}, the weight of pseudo labeled data $w_p$ $\in$ \{0.1, 0.3, 0.5, 0.7, 0.9\}, and the threshold for pseudo labeled data selection $\in$ \{0.01, 0.015, 0.02, 0.025, 0.03, 0.035, 0.04\}. 

The search results indicate that the optimal hyper-parameters of ssCDL on Nl27k are as follows: the standard deviation of Gaussian distribution $\sigma=0.6$, the weight of pseudo labeled data $w_p=0.7$, the MSE weight $\beta=1$, the learning rate $\alpha=0.001$, the margin $\gamma=0.1$, the threshold for pseudo labeled data selection: 0.03, the batch size: 4096, and the embedding size: 128. The optimal hyper-parameters of ssCDL on CN15k are as follows:
the standard deviation of Gaussian distribution $\sigma=0.6$, the weight of pseudo labeled data $w_p=0.3$, the MSE weight $\beta=1$, the learning rate $\alpha=0.001$, the margin $\gamma=0.1$, the threshold for pseudo labeled data selection: 0.015, the batch size: 4096, and the embedding size: 512. We evaluated ssCDL on the validation set every ten epochs. The maximum epochs on NL27k and CN15k were empirically set to 500 and 300, respectively. 

% The search results indicate that the optimal batch size, embedding size and margin on NL27k are 4096, 128, and 0.1, respectively. The optimal batch size, embedding size and margin on CN15k are 4096, 512, and 0.1, respectively. 

\section{Parameter Sensitivity}
\label{app:PS}
\begin{figure*}
\centering
\begin{subfigure}[b]{0.32\textwidth}
        \includegraphics[width=\textwidth]{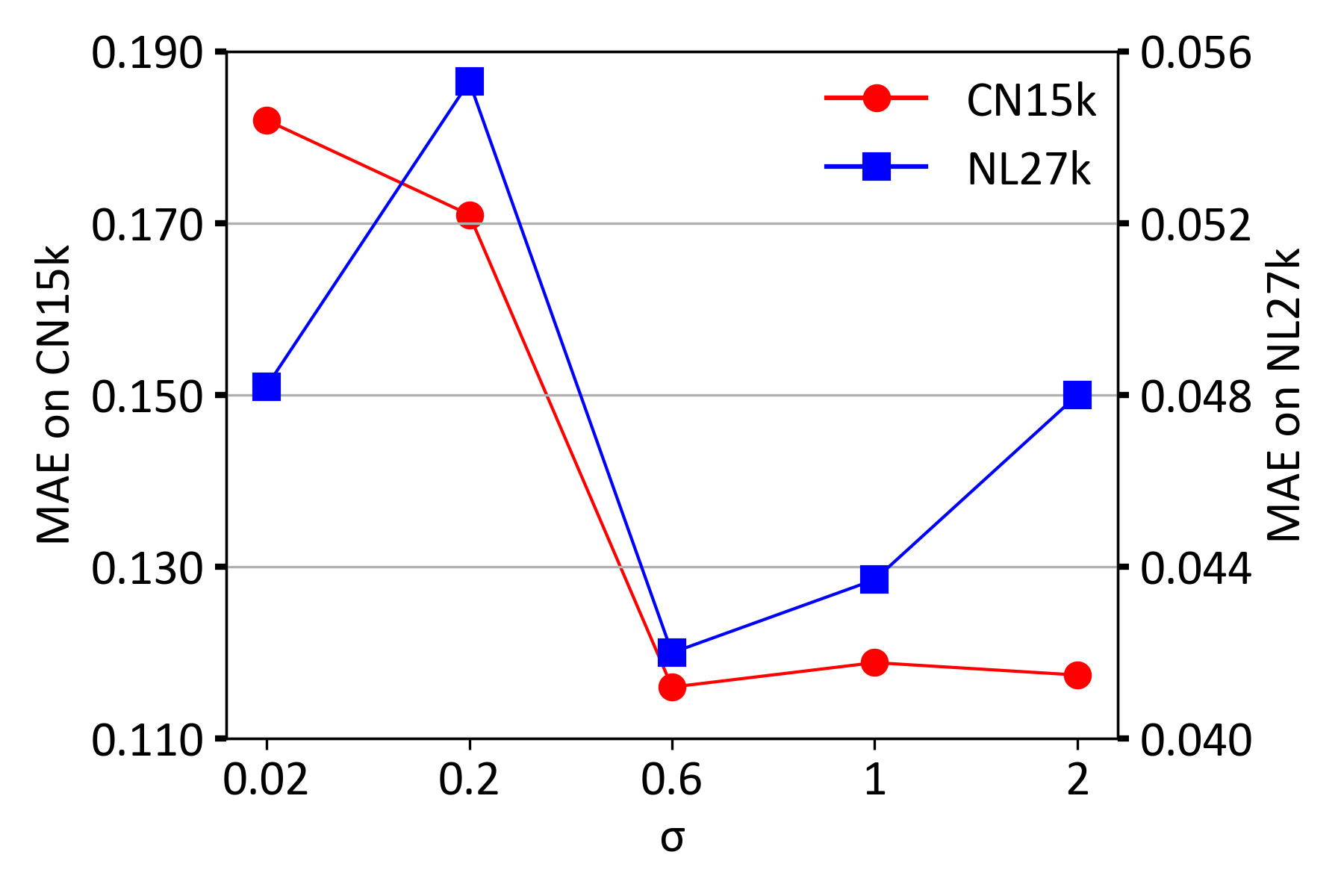}
        \caption{}
        \label{fig:1a1}
    \end{subfigure}
    \hfill % 填充空白
    \begin{subfigure}[b]{0.32\textwidth}
        \includegraphics[width=\textwidth]{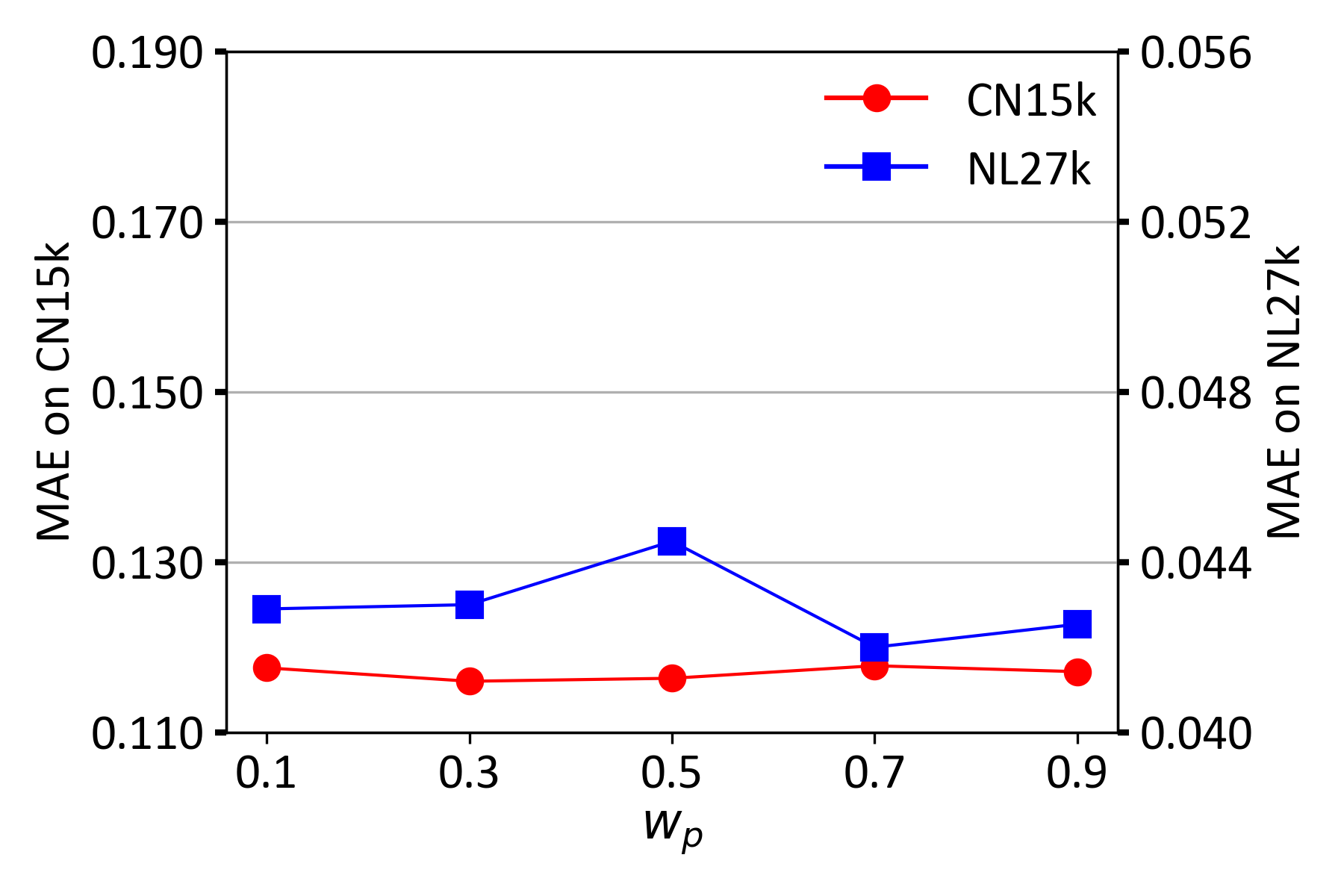}
        \caption{}
        \label{fig:1a2}
    \end{subfigure}
    \hfill % 填充空白
    \begin{subfigure}[b]{0.32\textwidth}
        \includegraphics[width=\textwidth]{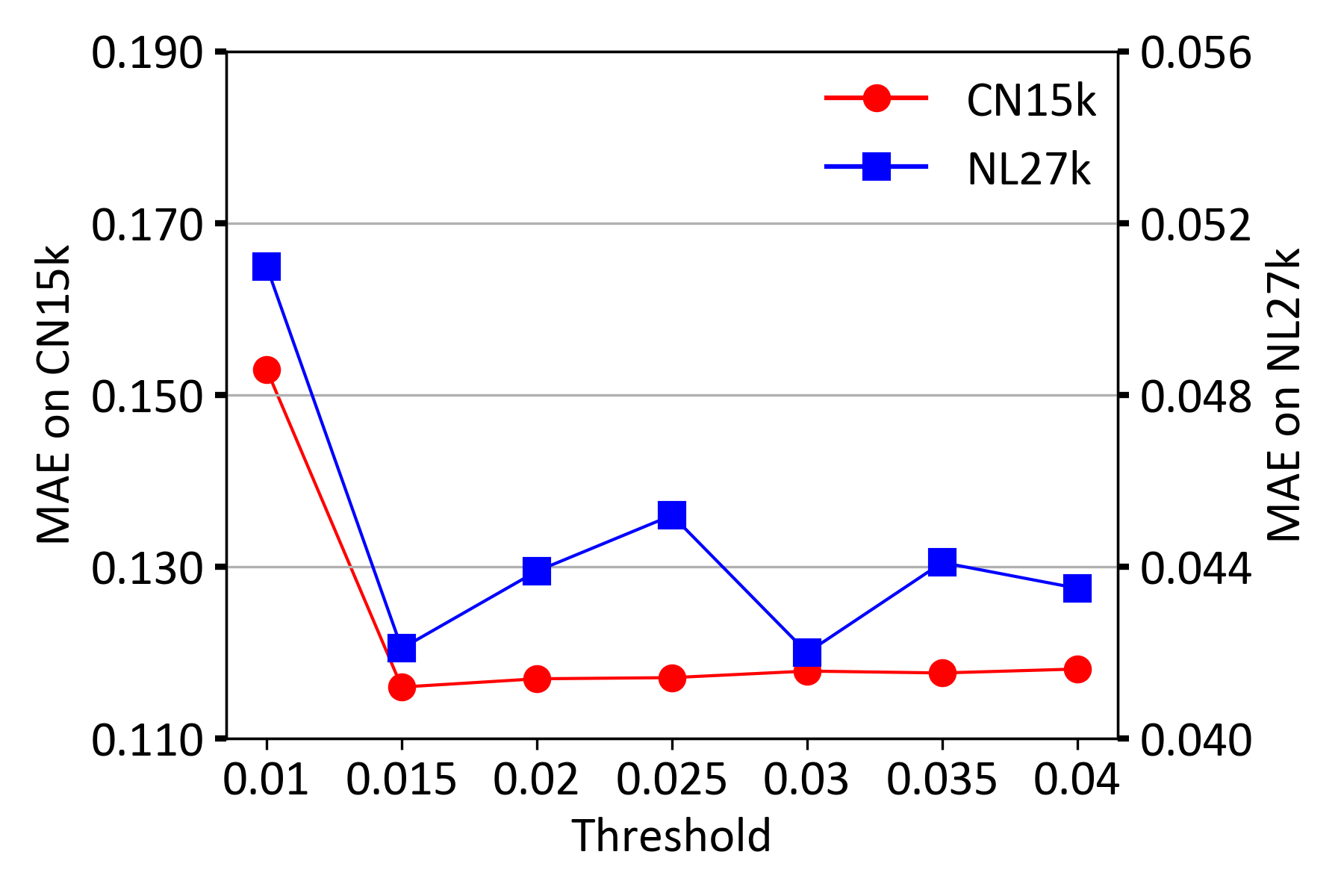}
        \caption{}
        \label{fig:1a3}
    \end{subfigure}
    \caption*{The impacts of hyper-parameters on confidence prediction. (a) (b) (c) show the impacts of $\sigma$, $w_p$, and the threshold for pseudo labeled data selection on NL27k and CN15k respectively.}
    \label{fig:1a}
    
    \vspace{0.5cm} % 增加两行图片之间的距离

    % 第二行，标记为1-b
    \begin{subfigure}[b]{0.32\textwidth}
        \includegraphics[width=\textwidth]{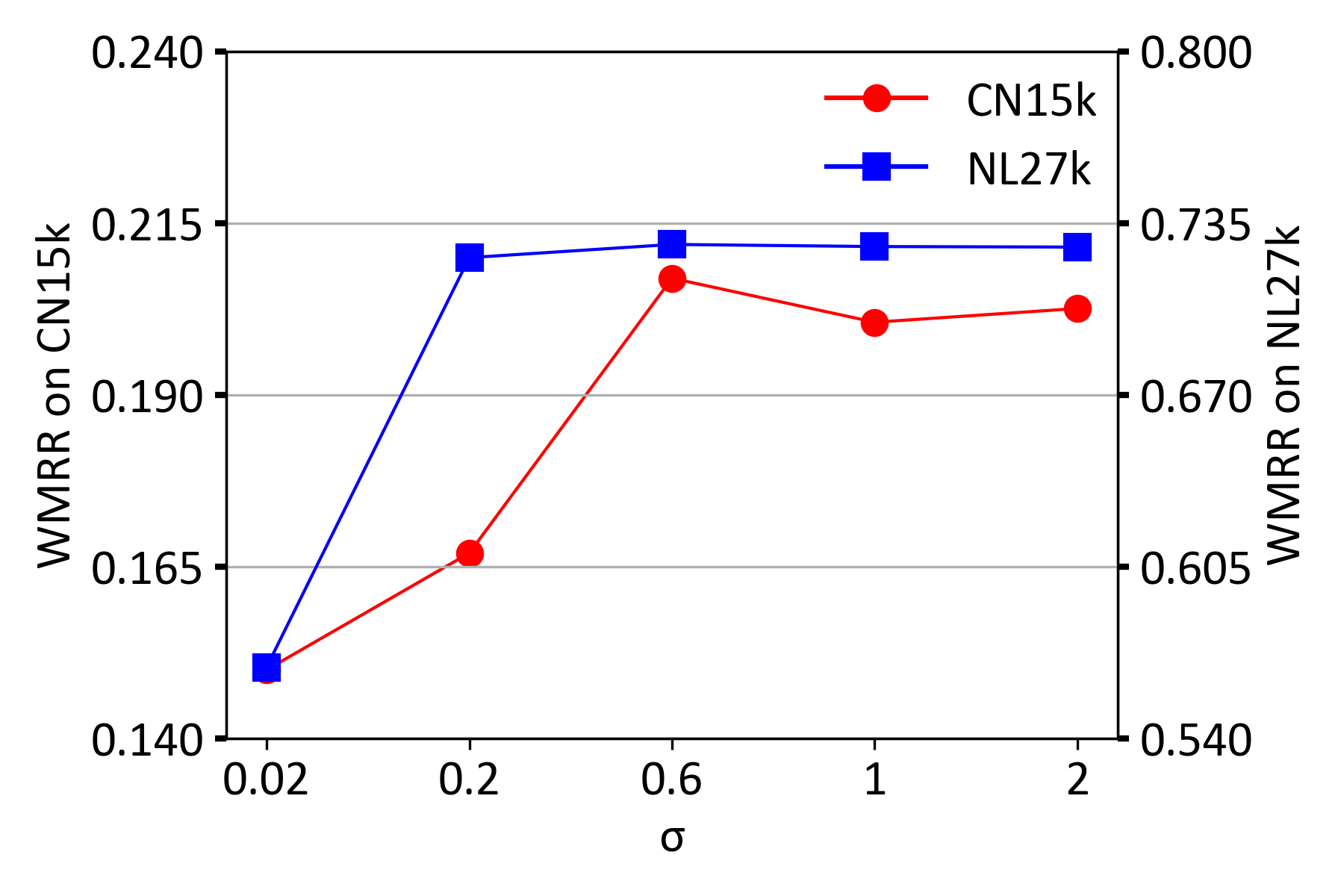}
        \caption{}
        \label{fig:1b1}
    \end{subfigure}
    \hfill % 填充空白
    \begin{subfigure}[b]{0.32\textwidth}
        \includegraphics[width=\textwidth]{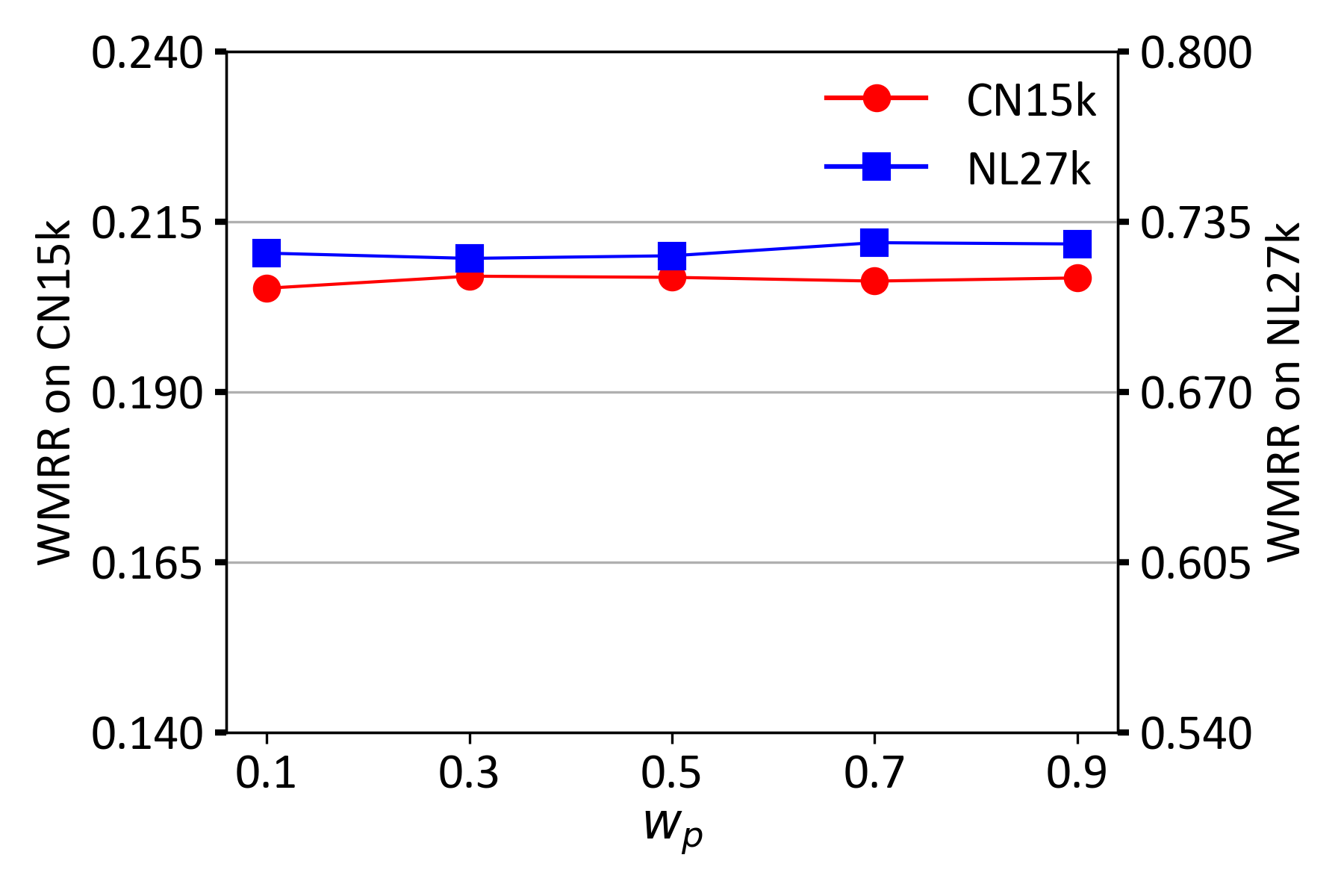}
        \caption{}
        \label{fig:1b2}
    \end{subfigure}
    \hfill % 填充空白
    \begin{subfigure}[b]{0.32\textwidth}
        \includegraphics[width=\textwidth]{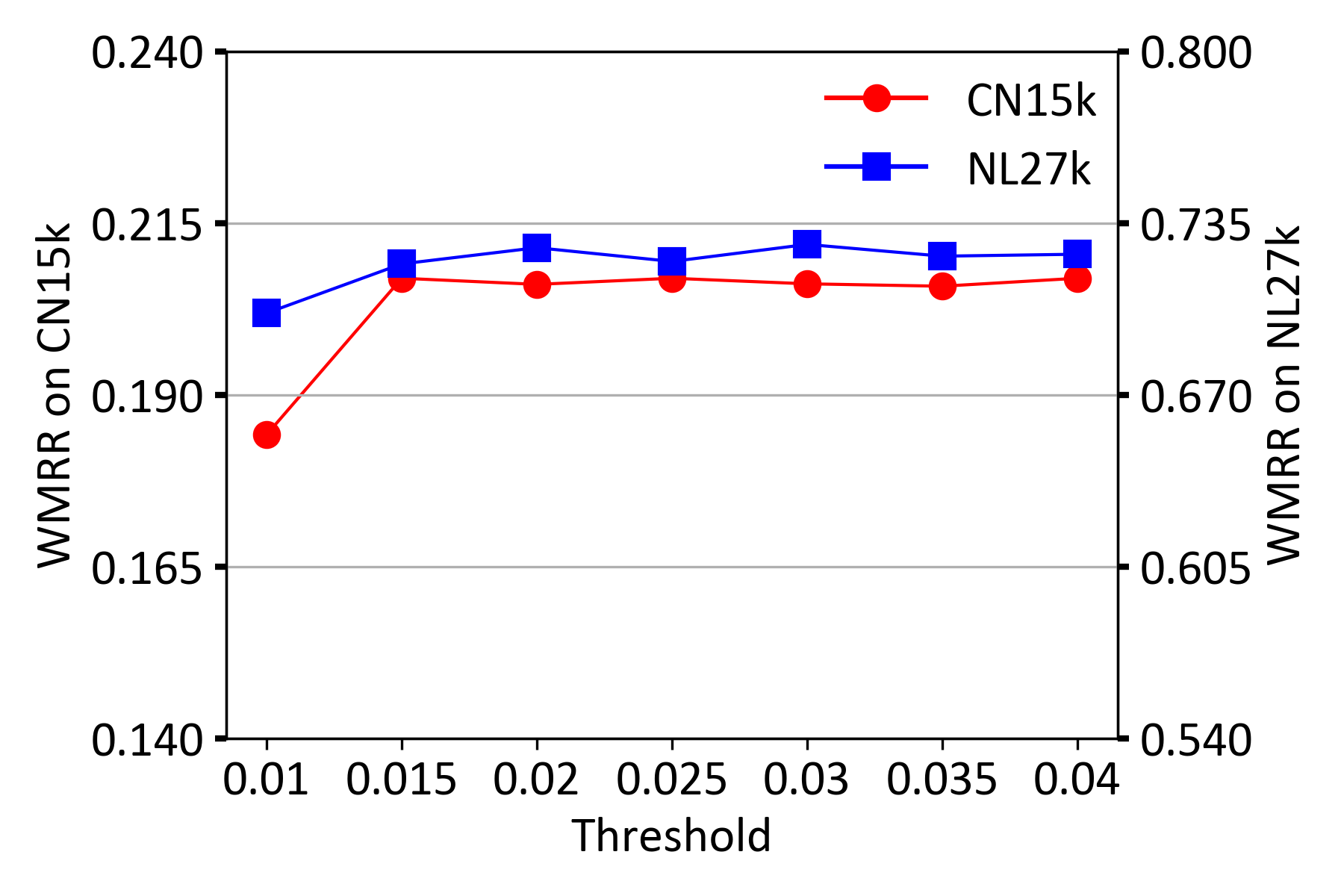}
        \caption{}
        \label{fig:1b3}
    \end{subfigure}
    \caption*{The impacts of hyper-parameters on link prediction. (d) (e) (f) show the impacts of $\sigma$, $w_p$, and the threshold for pseudo labeled data selection on NL27k and CN15k respectively.}
    \label{fig:1b}
    \caption{Sensitivity experiments on NL27k and CN15k.}

\label{fig:cross-column}
\end{figure*}

We conducted sensitivity analysis on three hyper-parameters, including the standard deviation of Gaussian distribution $\sigma$,  the weight of  pseudo labeled data $w_p$, and the threshold for pseudo labeled data selection. We utilized MAE and WMRR to evaluate the performance on confidence prediction and link prediction, respectively. Figure~\ref{fig:cross-column} shows the impacts of different hyper-parameters of ssCDL for confidence prediction and link prediction on both datasets.

% $\sigma$ is used to regulate the shape of the distribution. A larger $\sigma$ leads to a wider distribution, and both excessively wide or narrow distributions can negatively affect the performance of methods based on CDL. Therefore, it is crucial to find an appropriate $\sigma$ for ssCDL. In the context of meta self-training, $w_p$ and the threshold are used to regulate the influence of semi-supervised samples during the training process. $w_p$ is employed to control the weight of the loss of pseudo labeled data, while the threshold is used to control the quality of pseudo labeled data. If the impact of the pseudo labeled data is too small, the advantages of the meta self-training cannot be fully realized. On the other hand, if the influence of these samples is too large or their quality deteriorates, it may introduce noisy data. Therefore, it is important to analyze and select appropriate values for these two hyperparameters.

\spara{Impact of $\sigma$. }
As shown in Figure~\ref{fig:cross-column}(a) and \ref{fig:cross-column}(d), ssCDL achieves the optimal performance, then the effect deteriorates, as the value of $\sigma$ increases on both datasets. When the value of $\sigma$ is relatively small, the high description degree is mainly distributed in the confidence near the ground confidence, preventing the method from capturing the global features of UKGs, which leads to an inaccurate estimation of the confidence distribution. However, as $\sigma$ increases, the differences of confidence distributions generated by different ground confidences become small, which gradually weakens the ability of distinguishing between different confidences. ssCDL achieves the best results on both datasets when $\sigma=0.6$.

% Our experimental results indicate that $\sigma=0.6$ yields the best fitting performance for ssCDL on both datasets.
% On the other hand, when $\sigma$ is too large, the Gaussian distribution becomes excessively broad, causing the probability density function to flatten out, which results in a vague and inaccurate confidence distribution estimation that fails to reflect the real structure of UKG. 
% Figure XX visually demonstrates the issues that arise when $\sigma$ is either too small or too large.

\spara{Impact of $w_p$. }
For the hyper-parameter $w_p$, Figure \ref{fig:cross-column}(b) and \ref{fig:cross-column}(e) show that the performance of ssCDL improves at first and then decreases with the growing of $w_p$, which indicates that a reasonable balance between labeled data and pseudo labeled data is necessary to achieve the optimal performance of ssCDL. Our experimental findings indicate that 0.7 is the optimal $w_p$ on NL27k and 0.3 is the optimal $w_p$ on CN15k.
% The experimental results demonstrate that appropriately increasing the proportion of semi-supervised learning in the loss function can enhance the performance of ssCDL. However, an excessive weight assigned to pseudo labeled data may lead to erroneous judgments by the ssCDL, thereby significantly impacting the overall performance. Our experimental findings indicate that $w_p=0.3$ is a relatively optimal choice.

\spara{Impact of the threshold for pseudo labeled data selection.}
As shown in Figure \ref{fig:cross-column}(c) and \ref{fig:cross-column}(f), with the increase of the threshold, the performance of ssCDL also exhibits a trend of first increasing and then decreasing. A quite low threshold will cause that the pseudo label selection strategy to select noisy data, while a quite high threshold will limit the use of high-quality pseudo labeled data. ssCDL achieves the best results when the threshold is set as 0.03 on NL27k and 0.015 on CN15k, respectively.
% The influence of the threshold is similar to that of $w_p$, however, the threshold places greater emphasis on the quality of pseudo labeled data. A higher threshold imposes stricter requirements on the quality of pseudo labeled data generated by PCDG, which consequently leads to a reduction in the number of available pseudo labeled data. As shown in Figure~\ref{fig:1a3} and \ref{fig:1b3}, gradually increasing the threshold helps select higher-quality pseudo labeled data, thereby reducing noise. However, if the threshold is too high, the effectiveness of meta self-training may deteriorate. Our experiments ultimately reveal that setting the threshold as 0.8 is the optimal choice for both datasets.
\section{Limitation}
\label{app:LI}
Our current study has conducted experiments on CN15k. Although CN15k is a classic benchmark for UKG completion, since the most triples in ConceptNet are generally correct, there is no significant distinction in reliability between high-confidence and low-confidence triples. We will build better UKG completion benchmark datasets in the future.

\end{document}